\documentclass[journal]{IEEEtran}
\usepackage{times}
\usepackage{epsfig}
\usepackage{graphicx}
\usepackage{amsmath}
\usepackage{amssymb}
\usepackage{subfigure}
\usepackage{url}
\usepackage{calc}
\ifCLASSINFOpdf
\else
\fi
\begin{document}
%
\title{An All-in-One Network for Dehazing and Beyond}
%
%
%

\author{Boyi Li, Xiulian Peng,~\IEEEmembership{Member,~IEEE}, Zhangyang Wang,~\IEEEmembership{Member,~IEEE}, \\ Jizheng Xu,~\IEEEmembership{Senior Member,~IEEE}, Dan Feng,~\IEEEmembership{Member,~IEEE} 
\thanks{Boyi Li and Dan Feng are with the Wuhan National Laboratory for Optoelectronics, Huazhong University of Science and Technology, Wuhan, China. Email: boyilics@gmail.com, dfeng@hust.edu.cn.}
\thanks{Xiulian Peng and Jizheng Xu are with Microsoft Research, Beijing, China. Email: xipe@microsoft.com, jzxu@microsoft.com.}
\thanks{Zhangyang Wang is with the Department of Computer Science and Engineering, Texas A\&M University, USA. Email: atlaswang@tamu.edu.}
\thanks{The work was done when Boyi Li was an intern at Microsoft Research.}
}

%
%

\markboth{Journal of \LaTeX\ Class Files,~Vol.~14, No.~8, August~2015}%
{Shell \MakeLowercase{\textit{et al.}}: Bare Demo of IEEEtran.cls for IEEE Journals}
%



\maketitle

\begin{abstract}
   This paper proposes an image dehazing model built with a convolutional neural network (CNN), called All-in-One Dehazing Network (AOD-Net). It is designed based on a re-formulated atmospheric scattering model. Instead of estimating the transmission matrix and the atmospheric light separately as most previous models did, AOD-Net directly generates the clean image through a light-weight CNN. Such a novel end-to-end design makes it easy to embed AOD-Net into other deep models, e.g., Faster R-CNN, for improving high-level task performance on hazy images. Experimental results on both synthesized and natural hazy image datasets demonstrate our superior performance than the state-of-the-art in terms of PSNR, SSIM and the subjective visual quality. Furthermore, when concatenating AOD-Net with Faster R-CNN and training the joint pipeline from end to end, we witness a large improvement of the object detection performance on hazy images.
\end{abstract}


%
\IEEEpeerreviewmaketitle

\section{Introduction}

The existence of haze, due to the presence of aerosols such as dust, mist, and fumes, adds complicated noise to the images captured by cameras. It dramatically degrades the visibility of outdoor images, where contrasts are reduced and surface colors become faint. Moreover, a hazy image will put the effectiveness of many subsequent high-level computer vision tasks in jeopardy, such as object detection and recognition. The \textit{dehazing} algorithms have thus been widely considered, as a challenging instance of (ill-posed) image restoration and enhancement. Similar to other problems like image denoising and super-resolution \cite{tico2008multi, li2010multi}, earlier dehazing work~\cite{narasimhan2003contrast,schechner2001instant,treibitz2009polarization,kopf2008deep} assumed the availability of multiple images from the same scene. However, the haze removal from one single image has now gained the dominant popularity, since it is more practical for realistic settings \cite{fattal2008single}.  This paper focused on the problem of single image \textit{dehazing}.

\subsection{Prior Work}

As a prior knowledge to be exploited for dehazing, the hazy image generation follows a well-received physical model (see Section \ref{phy} for details).  Apart from estimating a global \textit{atmospheric light} magnitude, the key to achieve haze removal has been recognized to be the recovery of the \textit{transmission matrix}. \cite{fattal2008single} proposed a physically-grounded method by estimating the albedo of the scene. \cite{he2011single, tang2014investigating} discovered the effective dark channel prior (DCP) to more reliably calculate the transmission matrix, followed by a series of works \cite{kratz2009factorizing,nishino2012bayesian,tarel2012vision}. \cite{meng2013efficient} enforced the boundary constraint and contextual regularization for sharper restored images. An accelerated method for the automatic recovery of the atmospheric light was presented in~\cite{sulami2014automatic}. \cite{zhu2015fast} developed a color attenuation prior and created a linear model of scene depth for the hazy image, and then learned the model parameters in a supervised way.  \cite{li2015simultaneous} illustrate the method to jointly estimate scene depth and recover the clear latent image from a foggy video sequence. \cite{berman2016non} proposed an algorithm based on the non-local prior (haze-line), based on the assumption that each color cluster in the clear image becomes a haze-line in RGB space. All above methods hinge on the physical model and various sophisticated image statistics assumptions. However, since the estimation of physical parameters from a single image is often inaccurate, the dehazing performance of the above methods appears not always satisfactory. Lately, as Convolutional Neural Networks (CNNs) have witnessed prevailing success in computer vision tasks, they have been introduced to image dehazing as well. DehazeNet~\cite{cai2016dehazenet} proposed a trainable model to estimate the transmission matrix from a hazey image. ~\cite{ren2016single} further exploited a multi-scale CNN (MSCNN), that first generated a coarse-scale transmission matrix and later refined it.

\begin{figure*}[t]
	\begin{center}
		\subfigure[Comparison on PSNR]{
			\includegraphics[width=3.4in,height=1.8in]{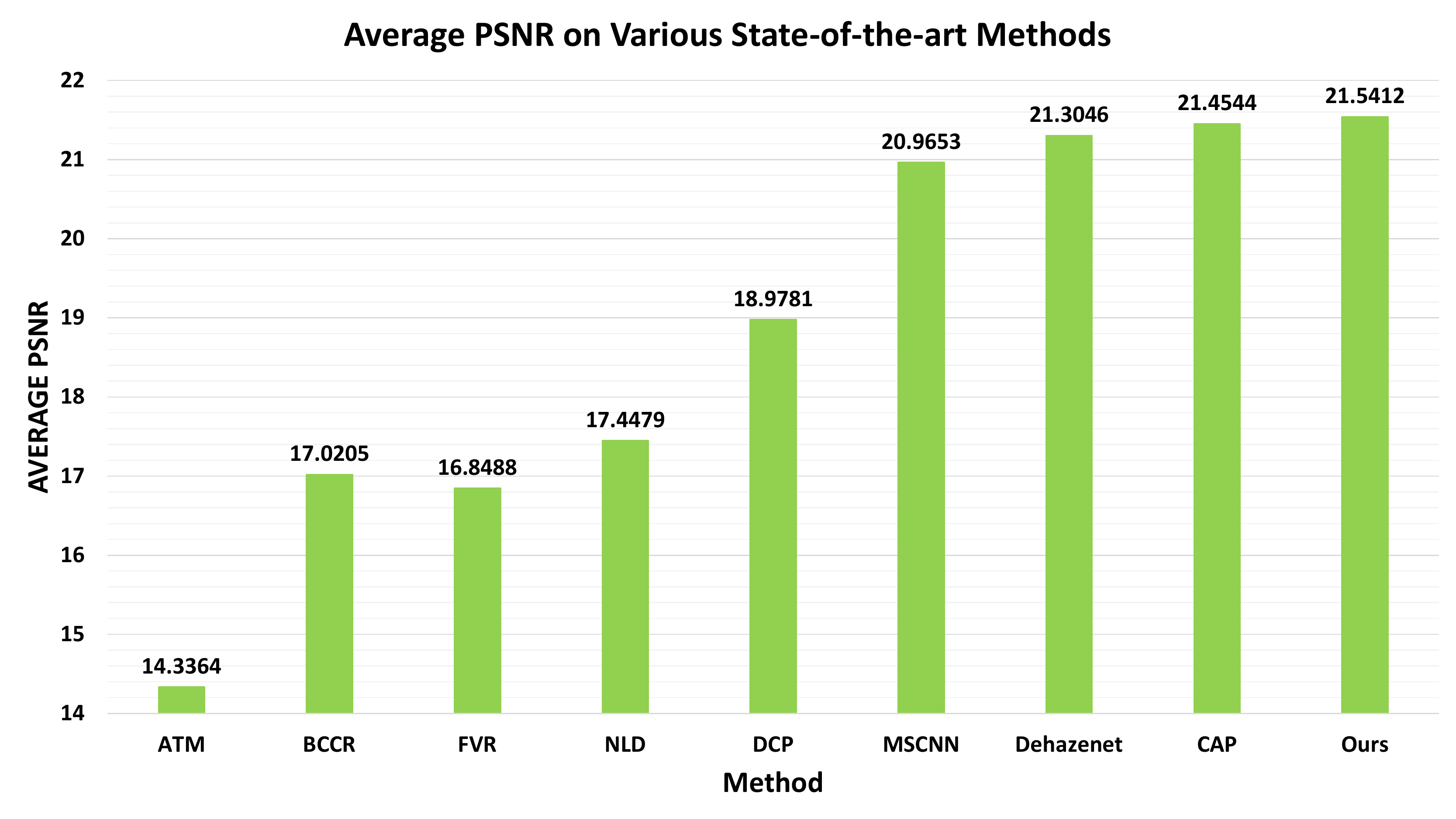}}\hfill
		\subfigure[Comparison on SSIM]{
			\includegraphics[width=3.4in,height=1.8in]{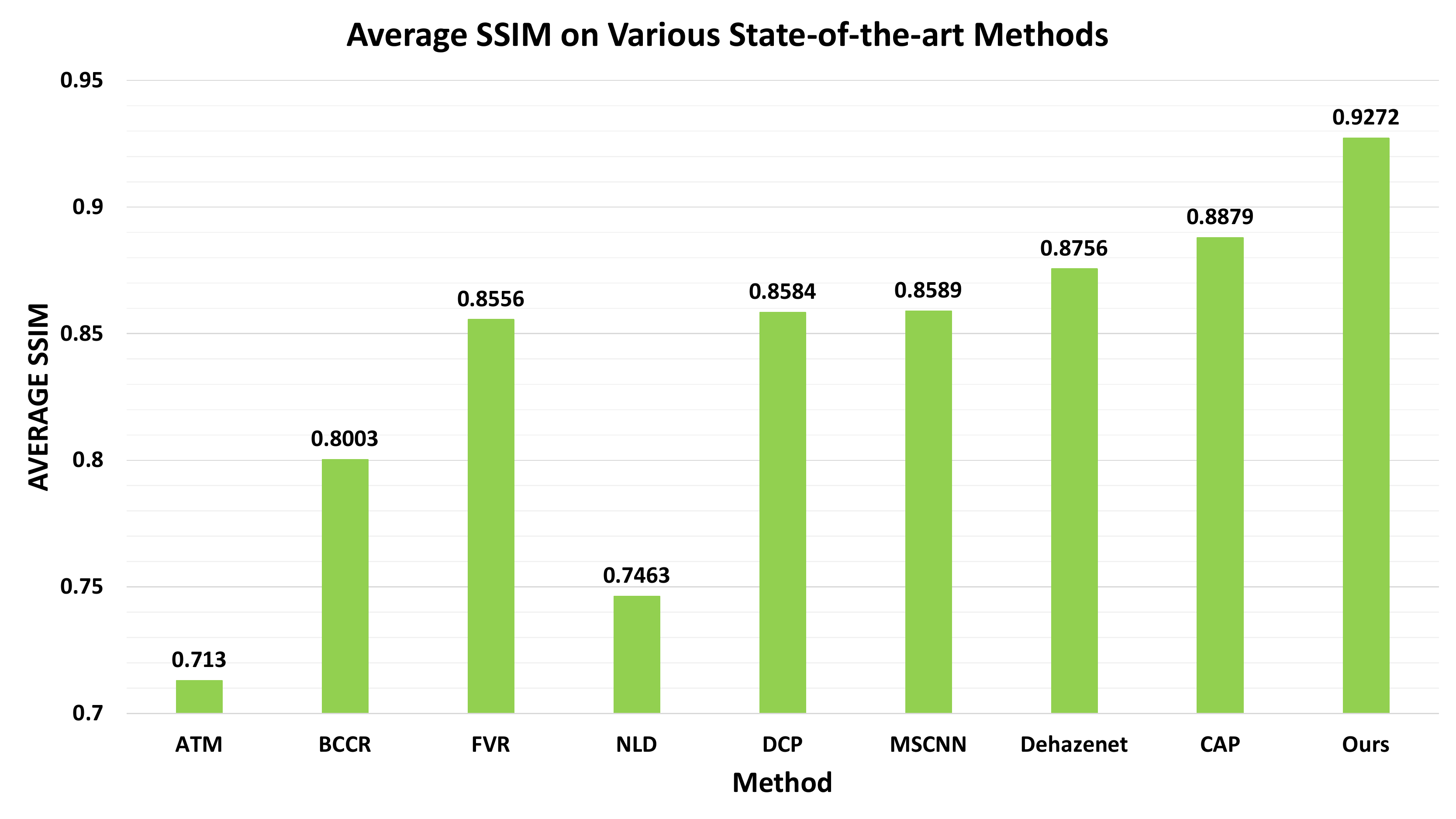}}
		\hspace{0.2in}
	\end{center}
	\caption{The PSNR and SSIM comparisons between AOD-Net and and several state-of-art methods on dehazing 800 synthetic images from Middlebury stereo database~\cite{scharstein2003high,scharstein2007learning,hirschmuller2007evaluation}. The results certify that AOD-Net presents more faithful reconstructions of clean images.}
	\label{fig:psnr1}
\end{figure*}

\begin{figure*}
	\centering
	\subfigure[Inputs]{
		\includegraphics[width=2.3in,height=1.8in]{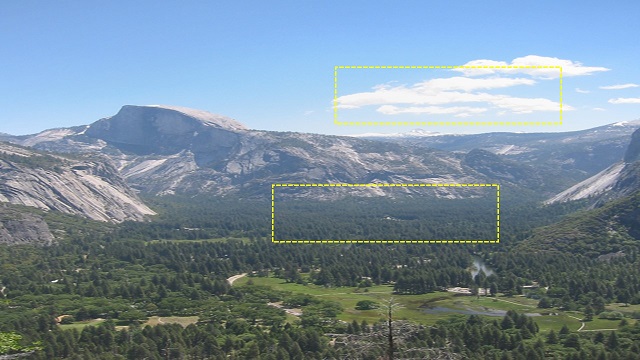}}
	\subfigure[DCP~\cite{he2011single}]{
		\includegraphics[width=2.3in,height=1.8in]{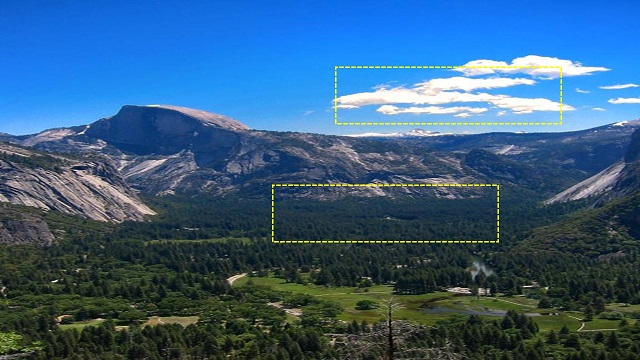}}
	\subfigure[CAP~\cite{zhu2015fast}]{
		\includegraphics[width=2.3in,height=1.8in]{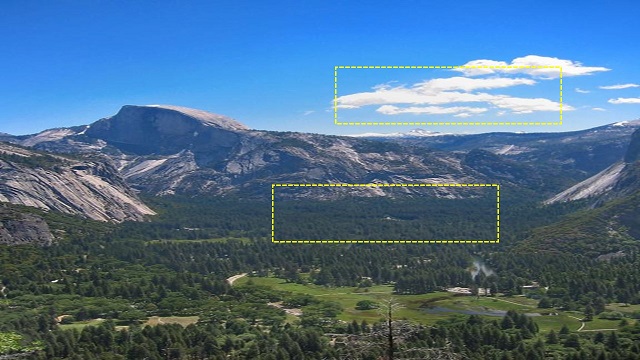}}
	\subfigure[DehazeNet~\cite{cai2016dehazenet}]{
		\includegraphics[width=2.3in,height=1.8in]{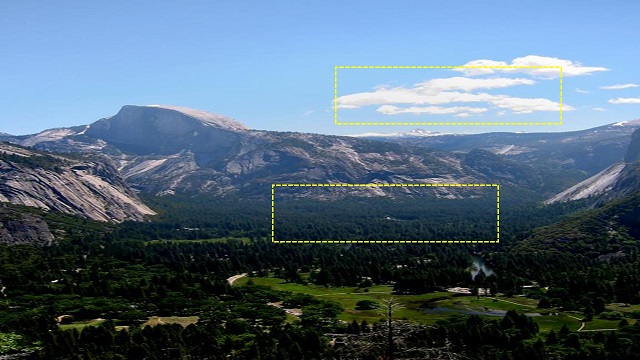}}
	\subfigure[MSCNN~\cite{ren2016single}]{
		\includegraphics[width=2.3in,height=1.8in]{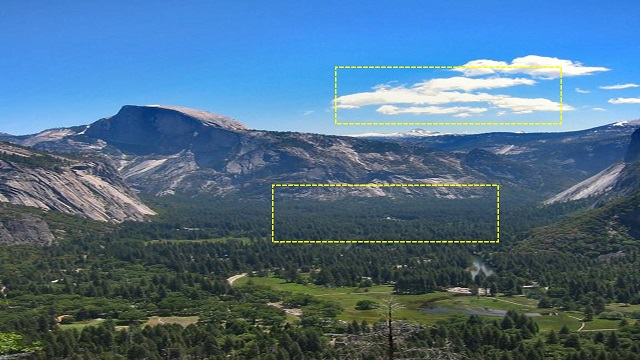}}
	\subfigure[AOD-Net]{
		\includegraphics[width=2.3in,height=1.8in]{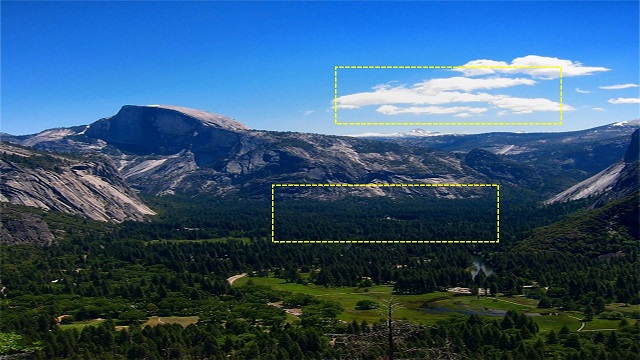}}
	\hspace{0.2in}
	\caption{Visual quality comparison between AOD-Net and several state-of-the-art methods on a natural hazy image. Please amplify figures to view the detail differences in bounded regions.}
	\label{fig:q1}
\end{figure*}

\subsection{Key Challenges and Bottlenecks}

\subsubsection{Absence of End-to-End Dehazing} Most deep learning approaches for image restoration and enhancement have fully embraced \textit{end-to-end modeling:} training a model to directly regress the clean image from the corrupted image. Examples include image denoising \cite{xie2012image}, deblurring \cite{schuler2016learning}, and super resolution \cite{wang2015self}. In comparison, \textit{there has been no end-to-end deep model for dehazing so far, that directly regresses a clean image from a hazy one}. While that might appear weird at the first glance, one needs to realize that haze essentially brings in non-uniform, \textit{signal-dependent noise}: the scene attenuation of a surface caused by haze is correlated with the physical distance between the surface the camera (i.e., the pixel depth). That is different from most image degradation models that assume \textit{signal-independent noise}, in which case all signals go through the same parameterized degradation process. Hence their restoration models  could be easily modeled with one static mapping function. The same is not directly applicable to dehazing: the degradation process varies by signals, and the restoration model has to be input-adaptive as well.

Existing methods share the same belief, that \textit{in order to recover a clean scene from haze, it is the key to estimate an accurate medium transmission map} \cite{berman2016non, cai2016dehazenet, ren2016single}. The atmospheric light is calculated separately by empirical rules, and the clean image is recovered based on the physical model. Albeit being intuitive, such a procedure does not directly measure or minimize the reconstruction distortions. The errors in the two separate steps for estimating transmission matrix and atmospheric light will accumulate and potentially amplify each other. As a result, the traditional separate pipeline gives rise to the sub-optimal image restoration quality.  

\subsubsection{Missing Link with High-Level Vision Tasks}

Currently, dehazing models rely on two sets of evaluation criteria: (1) for \textit{syntehtic hazy images} where their groundtruth clean images are known, PSNR and SSIM are typically computed to measure the restoration fidelity; (2) for \textit{real natural hazy images} with unknown groundtruth, the only available comparison of dehazing results is on the subjective visual quality. However, unlike image denoising and super resolution results whose suppression effects of visual artifacts are visible (e.g., on textures and edges), the visual differences between state-of-the-art dehazing models \cite{berman2016non, cai2016dehazenet, ren2016single} typically manifest in the global illumination and tone, and are often too subtle to tell. 

General image restoration and enhancement, known as part of low-level vision tasks, are usually thought as the pre-processing step for mid-level and high-level vision tasks. It has been known that the performance of high-level computer vision tasks, such as object detection and recognition, will deteriorate in the presence of various degradations, and is then largely affected by the quality of image restoration and enhancement. However, up to our best knowledge, there has been no exploration to correlate the dehazing algorithms and results with the high-level vision task performance.

\subsection{Main Contributions}

In this paper, we propose the \textit{All-in-One Dehazing Network} \textbf{(AOD-Net)}, a CNN-based dehazing model with two key innovations in response to the above two challenges:
\begin{itemize}
\item We are the first to propose an end-to-end trainable dehazing model, that directly produces the clean image from a hazy image, rather than relying on than \textit{any separate and intermediate parameter estimation step}\footnote{\cite{cai2016dehazenet} performed end-to-end learning from the hazy image to the transmission matrix, which is different from what we defined above.}. . 
AOD-Net is designed based on a re-formulated atmospheric scattering model, thus preserving the same physical ground as existing works \cite{cai2016dehazenet, ren2016single}. However, it is built with our different belief that \textit{the physical model could be formulated in a ``more end-to-end'' fashion, with all its parameters estimated in one unified model}.  

\item We are the first to quantitatively study how the dehazing quality could affect the subsequent high-level vision task, which serves as a new objective criterion for comparing dehazing results. Moreover, AOD-Net can be seamlessly embedded with other deep models, to constitute one pipeline that performs high-level tasks on hazy images, with an implicit dehazing process. Thanks to our unique all-in-one design, such a pipeline can be jointly tuned from end to end for improving performance further, which is infeasible if replacing AOD-Net with other deep hehazing alternatives~\cite{cai2016dehazenet, ren2016single}. 
\end{itemize}
AOD-Net is trained on synthetic hazy images, and tested on both synthetic and real natural images. Experiments demonstrate the superiority of AOD-Net over several state-of-the-art methods, in terms of not only PSNR and SSIM (see Figure~\ref{fig:psnr1}), but also visual quality (see Figure~\ref{fig:q1}). As a lightweight and efficient model, AOD-Net costs as low as 0.026 second to process one $ 480\times640 $ image with a single GPU. When concatenated with Faster R-CNN~\cite{NIPS2015_5638}, AOD-Net notably outperforms other dehazing models in improving the object detection performance over hazy images, and the performance margin is boosted even more when we jointly tune the pipeline of AOD-Net and Faster R-CNN from end to end. 

This paper is extended from a previous conference version \cite{AOD-Net}\footnote{The conference paper and codes are available at:  \url{https://sites.google.com/site/boyilics/website-builder/project-page}}. The most notable improvement of this current paper lies in Section \ref{detection}, where we present an in-depth discussion on evaluating and enhancing dehazing on object detection, and introduce the joint training part with abundant details and analysis. We also provide a more detailed and thorough analysis on the architecture of AOD-Net (e.g. Section \ref{comparemore}).  Besides, we have included more extensive comparison results.

\section{AOD-Net: The All-In-One Dehazing Model}

In this section, the proposed AOD-Net is explained. We first introduce the transformed atmospheric scattering model, based on which the AOD-Net is designed. The architecture of AOD-Net is then described in detail. 


\subsection{Physical Model and Transformed Formula}
\label{phy} 

The \textit{atmospheric scattering model} has been the classical description for the hazy image generation~\cite{mccartney1976optics,narasimhan2000chromatic,narasimhan2002vision}:
\begin{equation}\label{e1}
I\left( x\right) =J\left( x\right) t\left( x\right) +A\left( 1-t\left( x\right) \right) ,
\end{equation}
where $ I\left( x\right)  $ is observed hazy image, $ J\left( x\right)  $ is the scene radiance (i.e., the ideal ``clean image'') to be recovered. There are two critical parameters: $ A $ denotes the global atmospheric light, and $ t\left( x\right) $ is the transmission matrix defined as:
\begin{equation}\label{e}
t\left( x\right) =e^{-\beta d\left( x\right)},
\end{equation}
where $ \beta $ is the scattering coefficient of the atmosphere, and $ d\left( x\right) $ is the distance between the object and the camera. 
We can re-write the model \eqref{e1} for the clean image as the output:
\begin{equation}\label{e2}
J\left( x\right) =\dfrac{1}{t\left( x\right)} I\left( x\right)-A\dfrac{1}{t\left( x\right)}+A.
\end{equation}

\begin{figure}
	\centering
	\subfigure[Inputs]{
		\includegraphics[width=1in,height=2.4in]{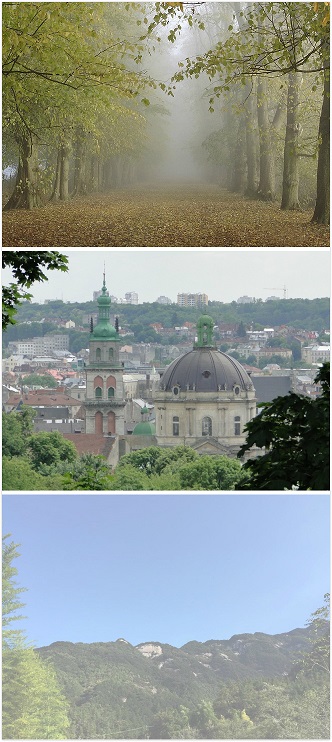}}
	\subfigure[AOD-Net using~\eqref{e3}]{
		\includegraphics[width=1in,height=2.4in]{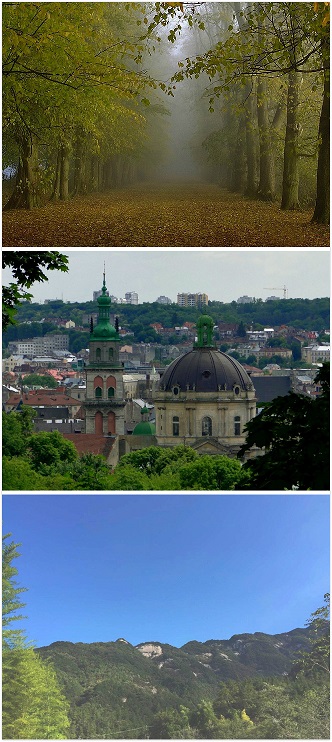}}
	\subfigure[Baseline using~\eqref{e2}]{
		\includegraphics[width=1in,height=2.4in]{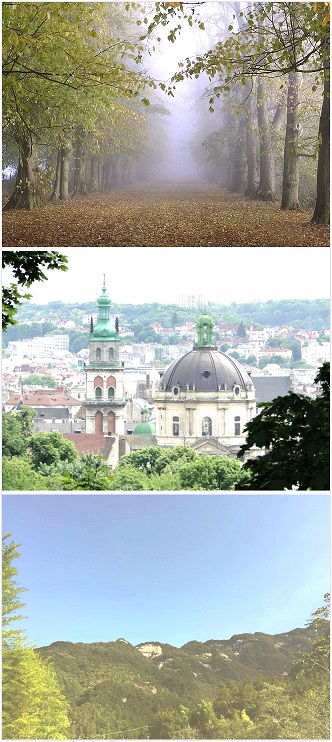}}
	\caption{Visual comparison between AOD-Net using~\eqref{e3}, and the naive baseline using~\eqref{e2}. The images are from the Challenging Real Photos in Section \ref{dehaze}.}
	\label{fig:light_A}
\end{figure}

Existing works such as~\cite{ren2016single, cai2016dehazenet} follows the identical three-step procedure: 1) estimating the transmission matrix $t\left( x\right)$ from the hazy image $I\left( x\right)$ using sophisticated deep models; 2) estimating $A$ using some empirical methods; 3) estimating the clean image $J\left( x\right)$ via \eqref{e2}. Such a procedure leads to a sub-optimal solution that does not directly minimize the image reconstruction errors. The separate estimation of $t\left( x\right)$ and $A$ will cause accumulated or even amplified errors, when combining them together to calculate \eqref{e2}.

 \begin{figure*}[tbp]
\centering
\begin{minipage}{0.55\textwidth}
\centering \subfigure[The diagram of AOD-Net] {
\includegraphics[width=\textwidth]{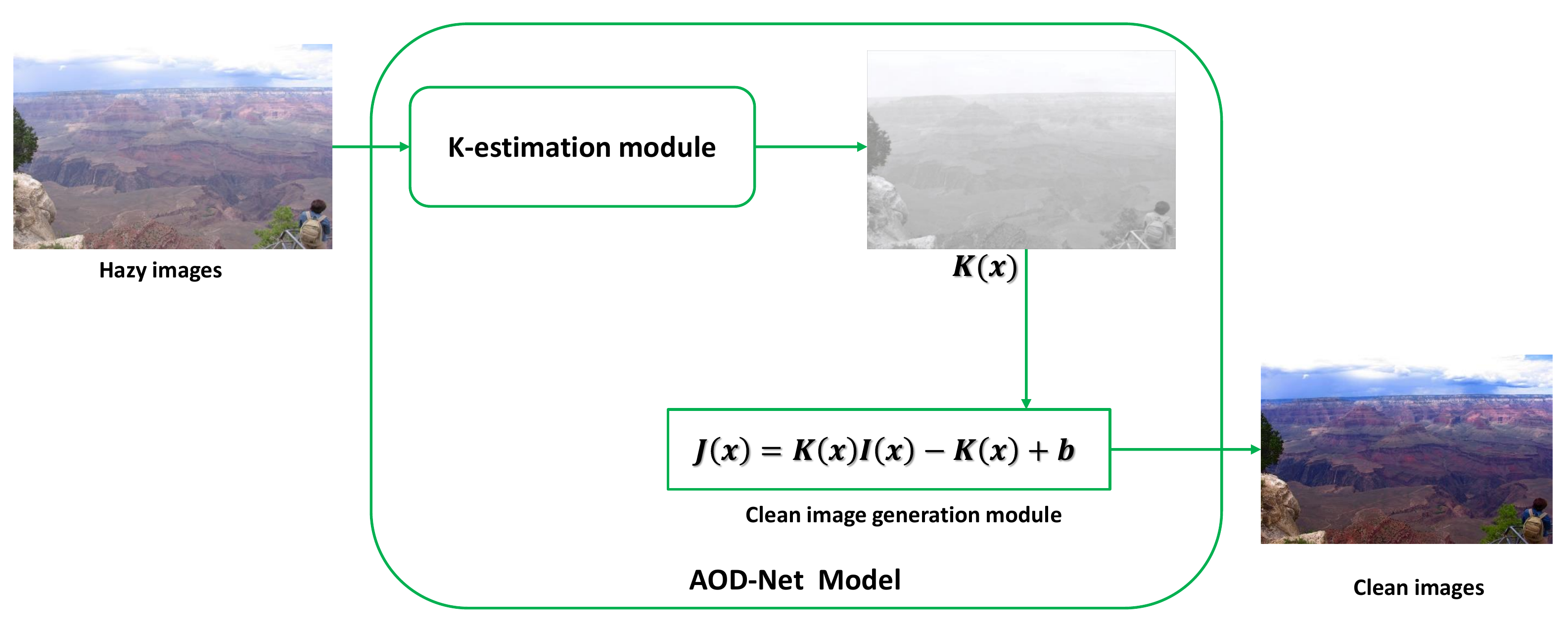}
}\end{minipage}
\begin{minipage}{0.40\textwidth}
\centering \subfigure[A closer look of the $K$-estimation module] {
\includegraphics[width=\textwidth]{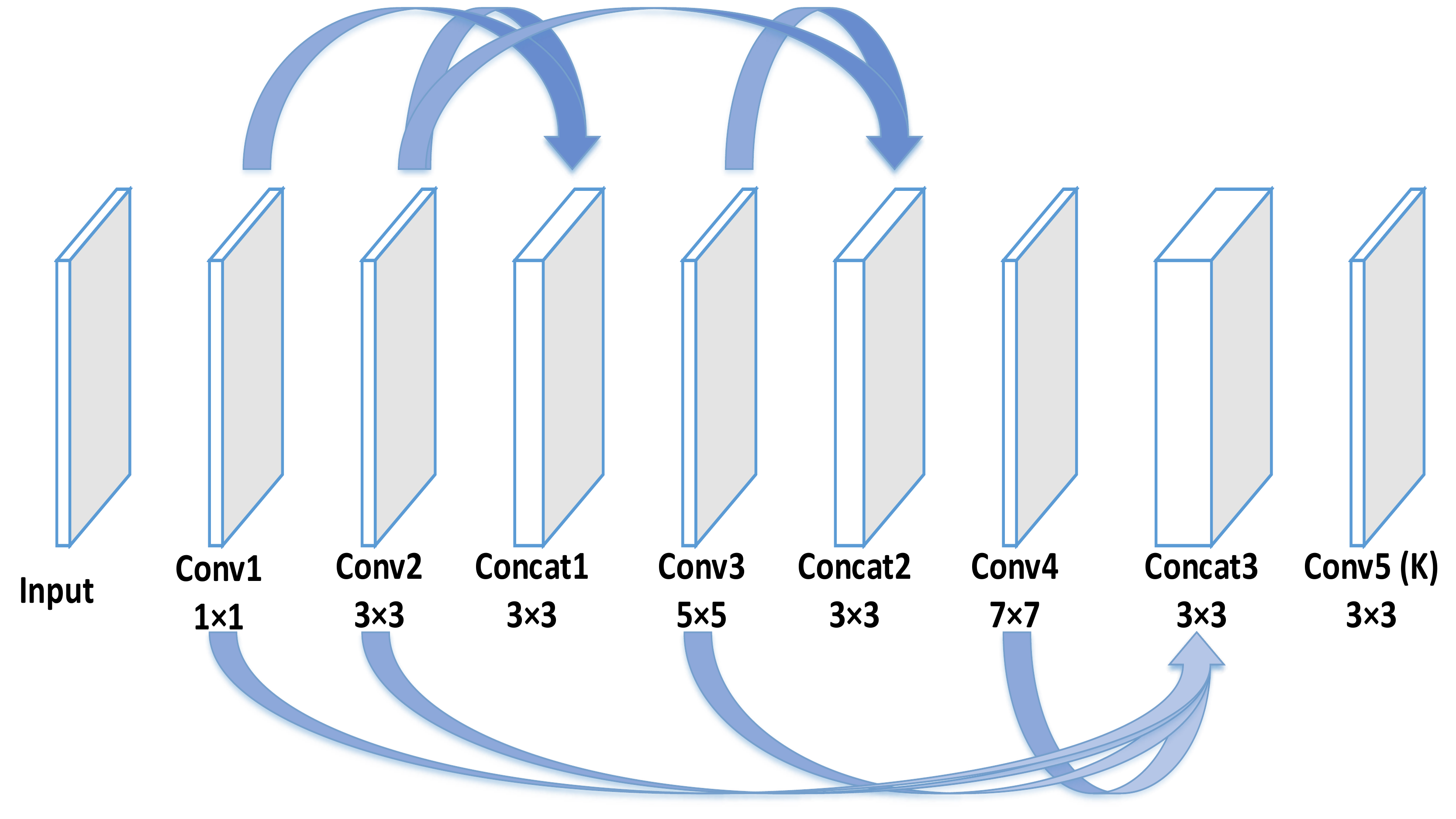}
}\end{minipage}
\caption{The network diagram and configuration of AOD-Net.}
\label{fig:struct}
\end{figure*}

Our core idea is to unify the two parameters $t\left( x\right)$ and $A$ into one formula, i.e. $K\left( x\right)$ in \eqref{e3}, and to directly minimize the pixel-domain reconstruction errors. To this end, the formula in \eqref{e2} is re-expressed as the \textit{Transformed Formula} below:
\begin{align}\label{e3}
\begin{split}
J\left( x\right) &=K\left( x\right) I\left( x\right) -K\left( x\right)+b, \text{where} \\
K\left( x\right) &= \dfrac{\frac{1}{t\left( x\right)}(I\left( x\right)-A)+(A-b)}{I\left( x\right)-1}.
\end{split}
\end{align}
In that way, both $\frac{1}{t\left( x\right)}$ and $A$ are integrated into the new variable $K\left( x\right)$. $b$ is the constant bias with the default value 1. Since $K\left( x\right)$ is dependent on $I\left( x\right)$, we then aim to build an \textit{input-adaptive} deep model, whose parameters will change with input hazy images, that minimizes the reconstruction error between the output $J\left( x\right)$ and the groundtruth clean image.

\subsection{Network Design}
The proposed AOD-Net consists of two modules, as illustrated in Figure~\ref{fig:struct} (a): a \textit{$K$-estimation module} that estimates $K\left( x\right)$ from the input $I\left( x\right)$, followed by a \textit{clean image generation module} that utilizes $K\left( x\right)$ as its input-adaptive parameters to estimate $J\left( x\right)$.





The $K$-estimation module is the critical component of AOD-Net, being responsible for estimating the depth and relative haze level. As depicted in Figure~\ref{fig:struct} (b), we use five convolutional layers, and form multi-scale features by fusing varied size filters. In~\cite{cai2016dehazenet}, parallel convolutions with varying filter sizes are used in the second layer. ~\cite{ren2016single} concatenated the coarse-scale network features with an intermediate layer of the fine-scale network. Inspired by them, the \textit{``concat1''} layer of AOD-Net concatenates features from the layers \textit{``conv1''} and \textit{``conv2''}. Similarly, \textit{``concat2''} concatenates those from \textit{``conv2''} and \textit{``conv3''}; \textit{``concat3''} concatenates those from \textit{``conv1''}, \textit{``conv2''}, \textit{``conv3''}, and \textit{``conv4''}. 
Such a multi-scale design captures features at different scales, and the intermediate connections also compensate for the information loss during convolutions. Notably, each convolutional layer of AOD-Net uses only three filters. As a result, AOD-Net is much light-weight, compared to existing deep methods, e.g.,~\cite{cai2016dehazenet, ren2016single}. Following the $K$-estimation module, the clean image generation module consists of an element-wise multiplication layer and several element-wise addition layers, in order to generate the recovered image via calculating (\ref{e3})

To justify why jointly learning $t\left( x\right)$ and $A$ in one is important, we compare a naive baseline that first estimates $A$ with the traditional method \cite{he2011single} and then learns $t\left( x\right)$ from \eqref{e2} using an end-to-end deep network by minimizing the reconstruction errors (see Section \ref{dehaze} for the synthetic settings). As observed in Figure~\ref{fig:light_A}, the baseline is found to overestimate $A$ and cause overexposure visual effects. AOD-Net clearly produces more realistic lighting conditions and structural details, since the joint estimation of  $\frac{1}{t\left( x\right)}$ and $A$ enables them to mutually refine each other. The inaccurate estimate of other hyperparameters (e.g., the gamma correction), can also be compromised and compensated in the all-in-one formulation.





\section{Evaluations on Dehazing}
\label{dehaze}
\subsection{Datasets and Implementation}
We create \textit{synthesized hazy images} by \eqref{e1}, using the ground-truth images with depth meta-data from the indoor NYU2 Depth Database~\cite{silberman2012indoor}. We set different atmospheric lights $A$, by choosing each channel uniformly between $[0.6,1.0]$, and select $ \beta\in\{0.4,0.6,0.8,1.0,1.2,1.4,1.6\}$. For the NYU2 database, we take 27, 256 images as the training set and 3,170 as the non-overlapping \textbf{TestSet A}. We also take the 800 full-size synthetic images from the Middlebury stereo database as the \textbf{TestSet B}. Besides, we test on \textit{natural hazy images} to evaluate our model generalization.

During the training process, the weights are initialized using Gaussian random variables. We utilize ReLU neuron as we found it more effective than the BReLU neuron proposed by \cite{cai2016dehazenet}, in our specific setting. The momentum and the decay parameter are set to 0.9 and 0.0001, respectively.  We use a batch size of 8 images($ 480 \times 640 $) and the learning rate is 0.001. We adopt the simple Mean Square Error (MSE) loss function, and are pleased to find that it boosts not only PSNR, but also SSIM as well as visual quality. 

The AOD-Net model takes around 10 training epochs to converge, and usually performs sufficiently well after 10 epochs. In this paper, we have trained the model for 40 epochs. It is also found helpful to clip the gradient to constrain the norm within $ [-0.1,0.1] $. The technique has been popular in stabilizing the recurrent network training~\cite{pascanu2013difficulty}.

\begin{figure*}
	\centering
	\includegraphics[width=6.8in,height=4in]{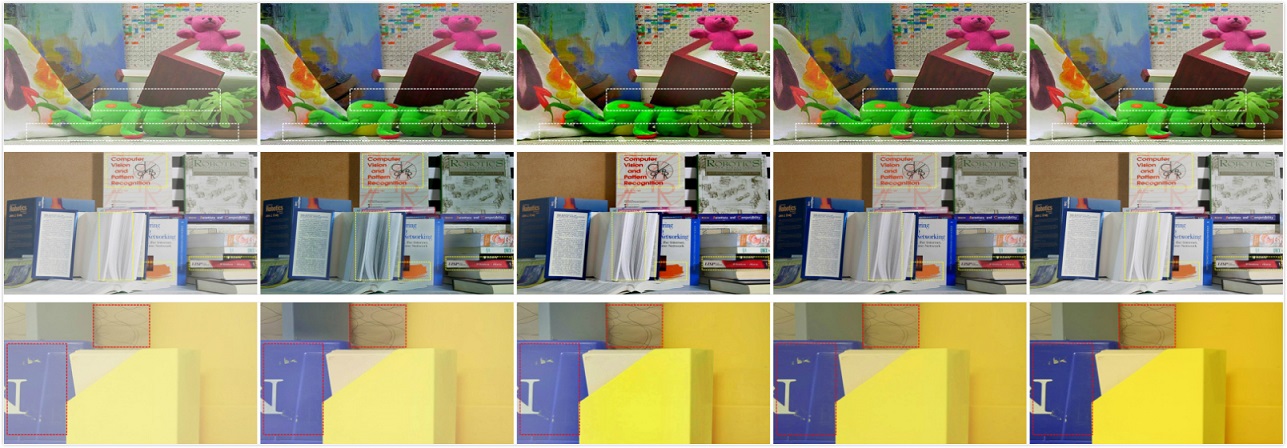}
	\hspace{0.2in}
	\caption{ Visual results on dehazing synthetic images. From left to right columns: hazy images, DehazeNet results~\cite{cai2016dehazenet}, MSCNN results~\cite{ren2016single}, AOD-Net results, and the groundtruth  images. Please amplify the figure to view the detail differences in bounded regions.}
	\label{fig:visualQ}
\end{figure*}

\subsection{Quantitative Results on Synthetic Images}
We compared the proposed model with several state-of-the-art dehazing methods: Fast Visibility Restoration (\textbf{FVR})~\cite{tarel2009fast}, Dark-Channel Prior (\textbf{DCP})~\cite{he2011single}, Boundary Constrained Context Regularization (\textbf{BCCR})~\cite{meng2013efficient}, Automatic Atmospheric Light Recovery (\textbf{ATM})~\cite{sulami2014automatic}, Color Attenuation Prior (\textbf{CAP})~\cite{zhu2015fast}, Non-local Image Dehazing (\textbf{NLD})~\cite{berman2016non,AirlightEstimation}, \textbf{DehazeNet}~\cite{cai2016dehazenet}, and \textbf{MSCNN}~\cite{ren2016single}. Among previous experiments, few quantitative results about the restoration quality were reported, due to the absence of haze-free ground-truth when testing on real hazy images. Our synthesized hazy images are accompanied with ground-truth images, enabling us to compare those dehazing results in terms of PSNR and SSIM. 

Tables~\ref{tab:psnrA} and \ref{tab:psnrB} display the average PSNR and SSIM results on TestSets A and B, respectively. Since AOD-Net is optimized from end to end under the MSE loss, it is not too surprising to see its higher PSNR performance than others. More appealing is the observation that AOD-Net gains even greater SSIM advantages over all competitors, even though SSIM is not directly referred to as an optimization criterion. As SSIM measures beyond pixel-wise errors and is well-known to more faithfully reflect the human perception, we become curious through which part of AOD-Net, such a consistent improvement is achieved. 

\begin{table*}
	\caption{Average PSNR and SSIM results on TestSet A.}
	\small
	\begin{center}
		\begin{tabular}{c|c|c|c|c|c|c|c|c|c}
			\hline
			Metrics&ATM~\cite{sulami2014automatic}&BCCR~\cite{meng2013efficient}&FVR~\cite{tarel2009fast}&NLD~\cite{berman2016non, AirlightEstimation}&DCP~\cite{he2011single}&MSCNN~\cite{ren2016single}&DehazeNet~\cite{cai2016dehazenet}&CAP~\cite{zhu2015fast}&\textbf{AOD-Net}\\
			\hline
			PSNR &14.1475&15.7606&16.0362&16.7653&18.5385&19.1116&18.9613&19.6364&\textbf{19.6954}\\
			\hline
			SSIM &0.7141&0.7711&0.7452&0.7356&0.8337&0.8295&0.7753&0.8374&\textbf{0.8478}\\
			\hline
		\end{tabular}
	\end{center}
	\label{tab:psnrA}
	\end{table*}
	
	\begin{table*}
	\begin{center}
		\caption{Average PSNR and SSIM results on TestSet B.}
		\small
		\begin{tabular}{c|c|c|c|c|c|c|c|c|c}
			\hline
			Metrics&ATM~\cite{sulami2014automatic}&BCCR~\cite{meng2013efficient}&FVR~\cite{tarel2009fast}&NLD~\cite{berman2016non,AirlightEstimation}&DCP~\cite{he2011single}&MSCNN~\cite{ren2016single}&DehazeNet~\cite{cai2016dehazenet}&CAP~\cite{zhu2015fast}&\textbf{AOD-Net}\\
			\hline
			PSNR &14.3364&17.0205&16.8488&17.4479&18.9781&20.9653&21.3046&21.4544&\textbf{21.5412}\\
			\hline
			SSIM &0.7130&0.8003&0.8556&0.7463&0.8584&0.8589&0.8756&0.8879&\textbf{0.9272}\\
			\hline
		\end{tabular}
	\end{center}
	\label{tab:psnrB}
\end{table*}

\begin{table*}
	\caption{Average MSE between the mean images of the dehazed images and the groundtruth images, on TestSet B.}
	\small
	\begin{center}
		\begin{tabular}{c|c|c|c|c|c|c|c|c|c}
			\hline
			Metrics&ATM~\cite{sulami2014automatic}&BCCR~\cite{meng2013efficient}&FVR~\cite{tarel2009fast}&NLD~\cite{berman2016non,AirlightEstimation}&DCP~\cite{he2011single}&MSCNN~\cite{ren2016single}&DehazeNet~\cite{cai2016dehazenet}&CAP~\cite{zhu2015fast}&\textbf{AOD-Net}\\
			\hline
			MSE &4794.40&917.20&849.23&2130.60 &664.30&329.97&424.90&356.68&\textbf{260.12}\\
			\hline
		\end{tabular}
	\end{center}	
	\label{tab:mean}
\end{table*}

\begin{figure*}
	\subfigure[Inputs]{
		\includegraphics[width=1.3in,height=4in]{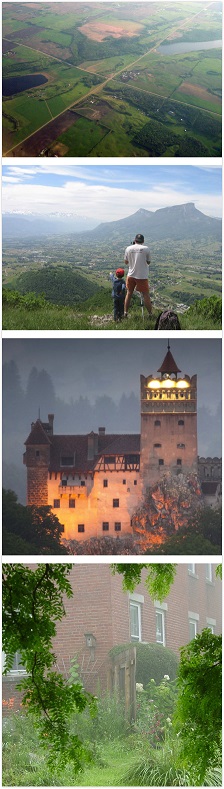}}
	\subfigure[FVR]{
		\includegraphics[width=1.3in,height=4in]{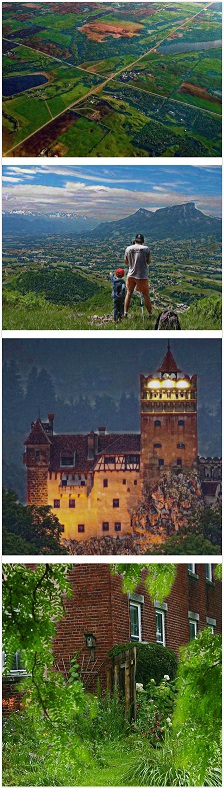}}
	\subfigure[DCP]{
		\includegraphics[width=1.3in,height=4in]{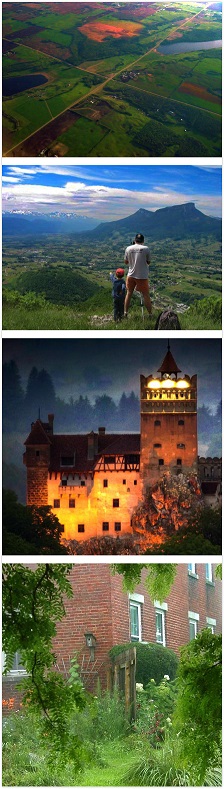}}
	\subfigure[BCCR]{
		\includegraphics[width=1.3in,height=4in]{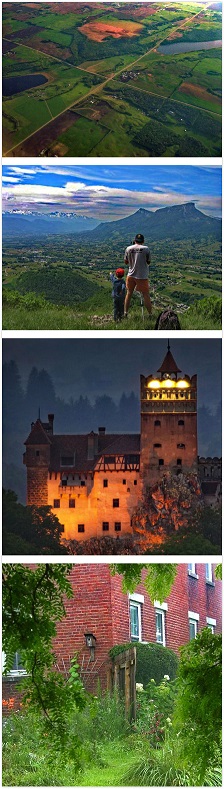}}
	\subfigure[ATM]{
		\includegraphics[width=1.3in,height=4in]{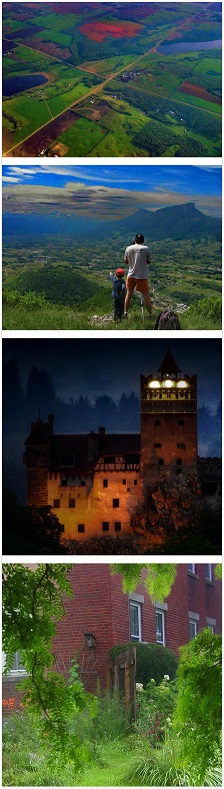}}\\
	\subfigure[CAP]{
		\includegraphics[width=1.3in,height=4in]{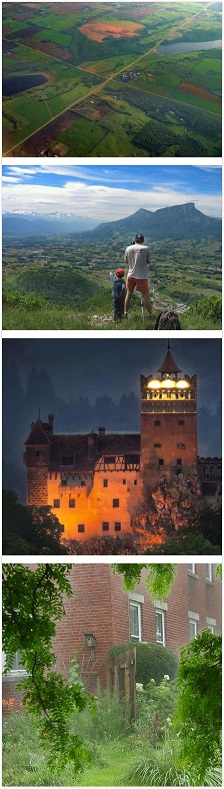}}
	\subfigure[NLD~\cite{berman2016non,AirlightEstimation}]{
		\includegraphics[width=1.3in,height=4in]{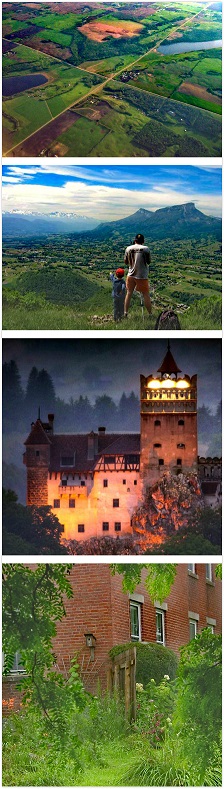}}
	\subfigure[DehazeNet~\cite{cai2016dehazenet}]{
		\includegraphics[width=1.3in,height=4in]{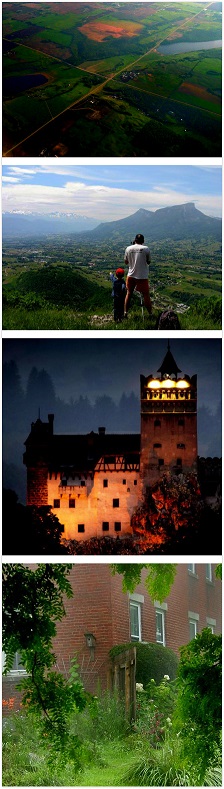}}
	\subfigure[MSCNN~\cite{ren2016single}]{
		\includegraphics[width=1.3in,height=4in]{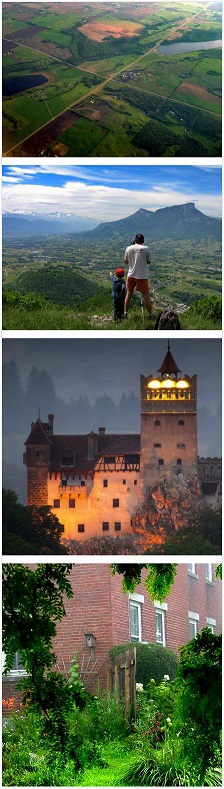}}
	\subfigure[AOD-Net]{
		\includegraphics[width=1.3in,height=4in]{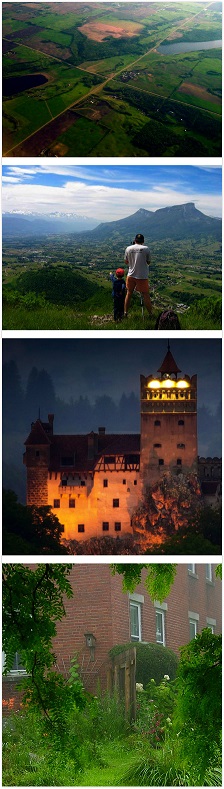}}
	\hspace{0.2in}
	\caption{Challenging natural images results compared with the state-of-the-art methods.}
	\label{fig:vq2}
\end{figure*}

We conduct the following investigation: each image in TestSet B is decomposed into the sum of a mean image and a residual image. The former is constructed by all pixel locations taking the same mean value (the average 3-channel vector across the image). It is easily justified that the MSE between the two images equals the MSE between their mean images added with that between two residual images. The mean image roughly corresponds to the global illumination and is related to $A$, while the residual concerns more the local structural variations and contrasts, etc. We observe that AOD-Net produces the similar residual MSE (averaged on TestSet B) to a few competitive methods such as DehazeNet and CAP. However, the MSEs of the mean parts of AOD-Net results are drastically lower than DehazeNet and CAP, as shown in Table~\ref{tab:mean}. Implied by that, AOD-Net could be more capable to correctly recover $A$ (global illumination), thanks to our joint parameter estimation scheme under an end-to-end reconstruction loss. Since the human eyes are certainly more sensitive to large changes in global illumination than to any local distortion, it is no wonder why the visual results of AOD-Net are also evidently better, while some other results often look unrealistically bright.

The above advantage also manifests in the \textit{illumination} ($l$) term of computing SSIM \cite{wang2004image}, and partially interprets our strong SSIM results. The other major source of SSIM gains seems to be from the \textit{contrast} ($c$) term. For examples, we randomly select five images for test, the mean of $contrast$ values of AOD-Net results on TestSetB is 0.9989, significantly higher than ATM (0.7281), BCCR (0.9574), FVR (0.9630), NLD(0.9250), DCP (0.9457) , MSCNN (0.9697), DehazeNet (0.9076), and CAP (0.9760).

\subsection{Qualitative Visual Results}

\paragraph{Synthetic Images} Figure~\ref{fig:visualQ} shows the dehazing results on synthetic images from TestSet A. We observe that AOD-Net results generally possess sharper contours and richer colors, and are more visually faithful to the ground-truth.

\paragraph{Challenging Natural Images} Although trained with synthesized by indoor images, ADO-Net is found to generalize well on outdoor images. We evaluate it against the state-of-art methods on a few natural image examples, that are significantly more challenging to dehaze than general outdoor images found by authors of \cite{he2011single,fattal2014dehazing,cai2016dehazenet}. The challenges lie the dominance of highly cluttered objects, fine textures, or illumination variations. As revealed by Figure~\ref{fig:vq2}, FVR suffers from overly-enhanced visual artifacts. DCP, BCCR, ATM, NLD, and MSCNN produce unrealistic color tones on one or several images, such as DCP, BCCR and ATM results on the second row (notice the sky color), or BCCR, NLD and MSCNN results on the fourth row (notice the stone color). CAP, DehazeNet, and AOD-Net have the most competitive visual results among all, with plausible details. Yet by a closer look, we still observe that CAP sometimes blurs image textures, and DehazeNet darkens some regions. AOD-Net recovers richer and more saturated colors (compare among third- and fourth-row results), while suppressing most artifacts.
\begin{figure*}
	\centering
	\subfigure[Input]{
		\includegraphics[width=1.1in,height=1.6in]{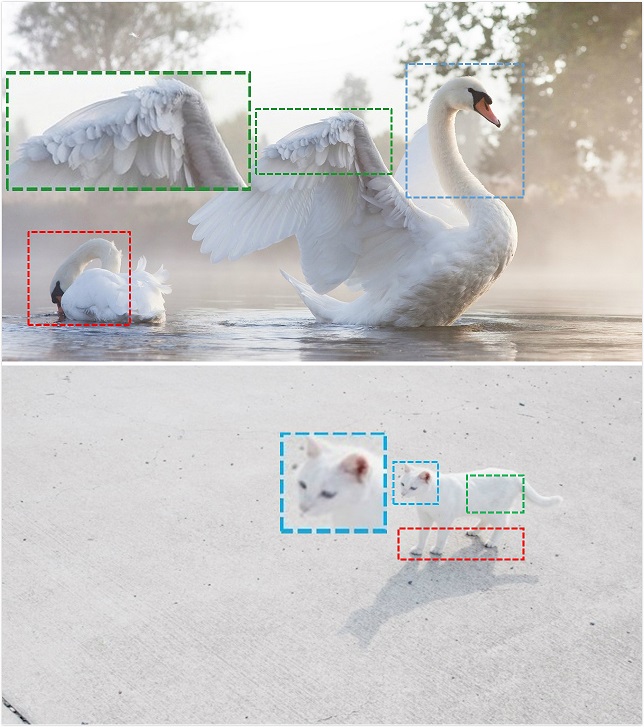}}
	\subfigure[DCP~\cite{he2011single}]{
		\includegraphics[width=1.1in,height=1.6in]{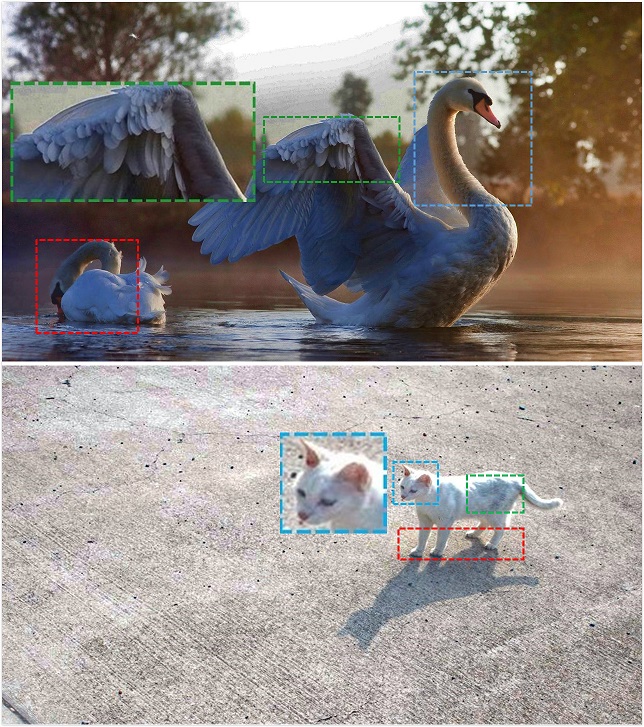}}
	\subfigure[CAP~\cite{zhu2015fast}]{
		\includegraphics[width=1.1in,height=1.6in]{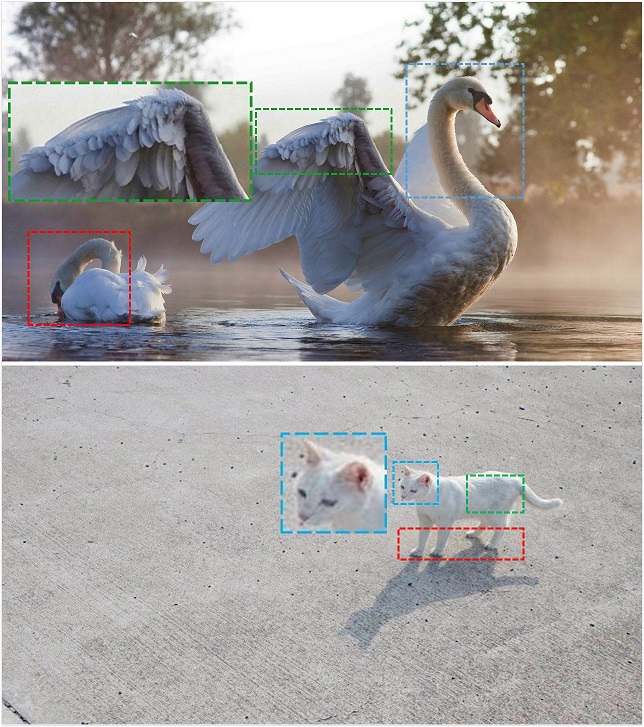}}
	\subfigure[DehazeNet~\cite{cai2016dehazenet}]{
		\includegraphics[width=1.1in,height=1.6in]{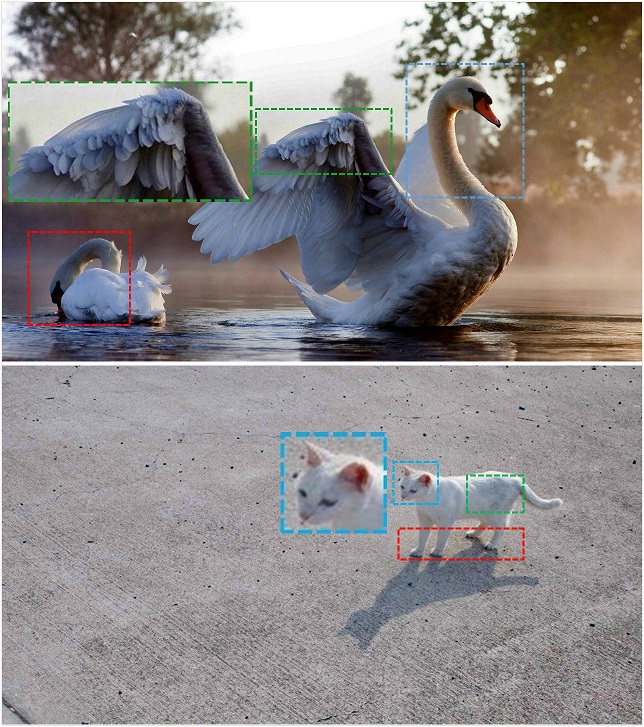}}
	\subfigure[MSCNN~\cite{ren2016single}]{
		\includegraphics[width=1.1in,height=1.6in]{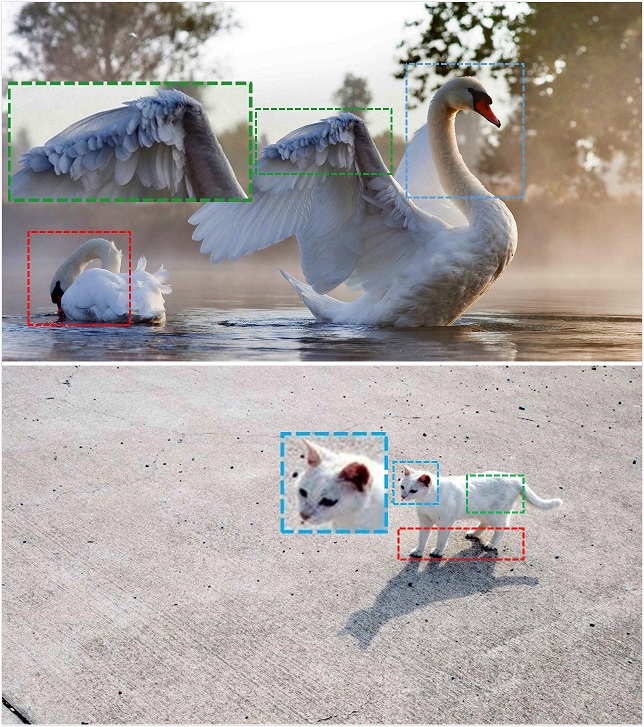}}
	\subfigure[AOD-Net]{
		\includegraphics[width=1.1in,height=1.6in]{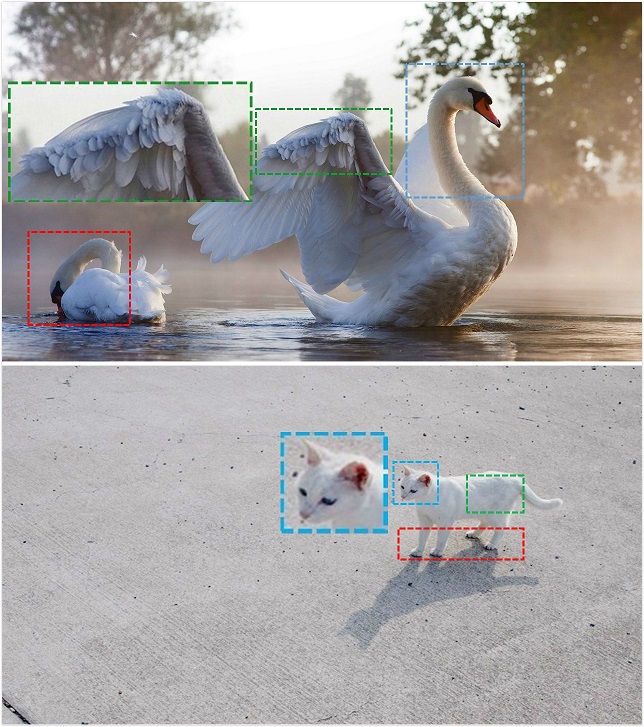}}
	\hspace{0.2in}
	\caption{White scenery image dehazing results. Please amplify figures to view the detail differences in bounded regions.}
	\label{fig:vq3}
\end{figure*}


\begin{figure}
	\begin{center}
		\includegraphics[width=3.5in,height=2.8in]{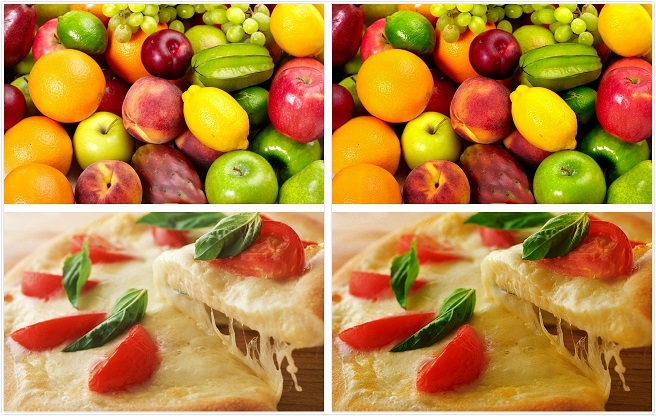}
	\end{center}
	\caption{Examples on impacts over haze-free images. Left column: haze-free images. Right column: outputs by AOD-Net.}
	\label{fig:clean_com}
\end{figure}

\begin{figure}
	\begin{center}
		\includegraphics[width=3.5in,height=3in]{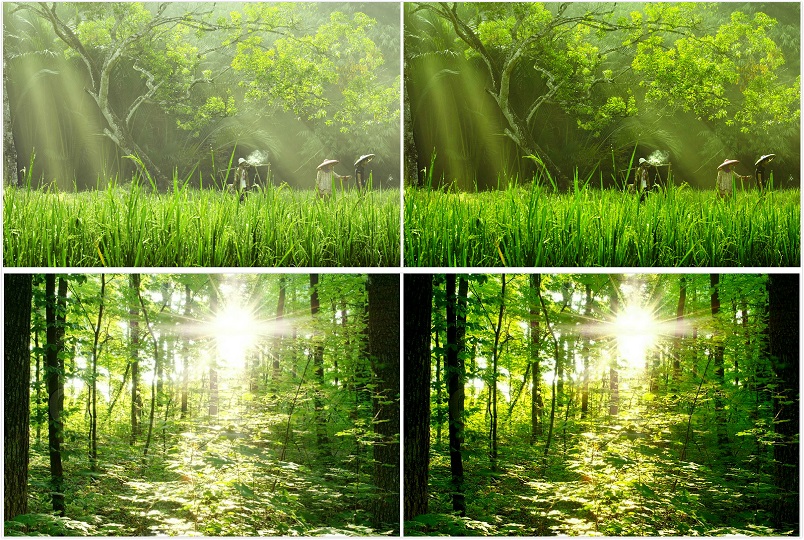}
	\end{center}
	\caption{Examples for anti-halation enhancement. Left column: real photos with halation. Right column: results by AOD-Net.}
	\label{fig:anti-halation}
\end{figure}

\paragraph{White Scenery Natural Images} White scenes or object has always been a major obstacle for haze removal. Many effective priors such as \cite{he2011single} fail on white objects since for objects of similar color to the atmospheric light, the transmission value is close to zero. DehazeNet~\cite{cai2016dehazenet}  and MSCNN~\cite{ren2016single} both rely on carefully-chosen filtering operations for post-processing, which improve their robustness to white objects but inevitably sacrifice more visual details.  

Although AOD-Net does not explicitly consider handling white scenes, our end-to-end optimization scheme seems to contribute stronger robustness here. Figure~\ref{fig:vq3} displays two hazy images of white scenes and their dehazing results by various methods. It is easy to notice the intolerable artifacts of DCP results, especially in the sky region of the first row. The problem is alleviated, but persists in CAP, DehazeNet and MSCNN results, while the AOD-Net is almost artifact-free. Moreover, CAP seems to blur the textural details on white objects, while MSCNN creates the opposite artifact of over-enhancement: see the cat head region for a comparison. AOD-Net manages to remove the haze, without introducing fake color tones or distorted object contours. 


\paragraph{Little Hurt on Haze-Free images} 
Although trained over hazy images, AOD-Net is verified to possess the highly desirable property, that it causes little negative impacts on the input image if it is haze-free. That endorses the robustness and effectiveness of our $K$-estimation module. Figure \ref{fig:clean_com} shows the results on two challenging clean images from Colorlines\cite{fattal2014dehazing}.

\paragraph{Image Anti-Halation} 
We try AOD-Net on another image enhancement task, called image anti-halation, \textit{without re-training}. Halation is a spreading of light beyond proper boundaries, forming an undesirable fog effect in the bright areas of photos. Being related to dehazing but following different physical models, the anti-halation results by AOD-Net are decent too: see Figure~\ref{fig:anti-halation} for a few examples.

\subsection{Effectiveness of Multi-Scale Features}
\label{comparemore}

In this section, we specifically analyze the usefulness of inter-layer concatenations of the $K$-estimation module, which combine multi-scale features from varied size filters. We conjecture that despite an empirical finding, the current concatenation way facilitates smooth feature transition from low-level to higher-level by consistently feeding several consecutive lower layers into the immediate next layer. For comparison purposes, we designed a baseline: \textit{``conv1 $ \rightarrow $ conv2 $ \rightarrow $ conv3 $ \rightarrow $ conv4 $ \rightarrow $ conv5($ K $)''}, that involves no inter-layer concatenation. For TestSet A, the average PSNR is 17.0517 dB and SSIM is 0.7688. For TestSet B, the average PSNR is 22.3359 dB and SSIM is 0.9032. Those results are generally inferior to AOD-Net (except PSNR on TestSet B gets marginally higher), especially both SSIM values suffer from significant drops.

\subsection{Running Time Comparison}

The light-weight structure of AOD-Net leads to faster dehazing. We select 50 images from TestSet A for all models to run, on the same machine (Intel(R) Core(TM) i7-6700 CPU@3.40GHz and 16GB memory), without GPU acceleration. The per-image average running time of all models are shown in Table~\ref{tab:runningtimeA}. Despite other slower Matlab implementations, it is fair to compare DehazeNet (Pycaffe version) and ours\cite{jia2014caffe}. The results illustrate the promising efficiency of AOD-Net, costing only 1/10 time of DehazeNet per image.

\begin{table}
	\caption{Comparison of average model running time (in seconds).}
	\begin{center}
		\begin{tabular}{l|c|l}
			\hline
			Image Size&$ 480\times640 $&Platform\\
			\hline
			ATM~\cite{sulami2014automatic}&35.19&Matlab\\
			
			DCP~\cite{he2011single}&18.38&Matlab\\
			
			FVR~\cite{tarel2009fast}&6.15&Matlab\\
			
			NLD~\cite{berman2016non,AirlightEstimation}&6.09&Matlab\\
			
			BCCR~\cite{meng2013efficient}&1.77&Matlab\\
			
			MSCNN~\cite{ren2016single}&1.70&Matlab\\
			
			CAP~\cite{zhu2015fast}&0.81&Matlab\\
			
			DehazeNet (Matlab)~\cite{cai2016dehazenet}&1.81&Matlab\\
			
			DehazeNet (Pycaffe)\footnote{https://github.com/zlinker/DehazeNet}~\cite{cai2016dehazenet}&5.09&Pycaffe\\
			
			\textbf{AOD-Net}&\textbf{0.65}&Pycaffe\\
			\hline
		\end{tabular}
	\end{center}
	
	\label{tab:runningtimeA}
\end{table}

%

	\begin{table*}
	\caption{The mAP Comparison on the synthetic hazy sets from PASCAL VOC 2007 (F: Faster R-CNN)}
	\small
	\begin{center}
		\begin{tabular}{ccccccccc}
			\hline
			Training Set&Test Set&Naive F&MSCNN+F&Dehazenet+F&DCP+F&AOD-Net+F&Retrained F&JAOD-Faster R-CNN\\
			\hline
			\textit{Heavy}&\textit{Heavy}&0.5155&0.5973&0.6138&0.6267&0.5794&0.6756&0.6819\\
			\hline
			\textit{Medium}&\textit{Medium}&0.6046&0.6451&0.6520&0.6622&0.6401&0.6920&0.6931\\
			\hline
			\textit{Light}&\textit{Light}&0.6410&0.6628&0.6628&0.6762&0.6701&0.6959&0.7004\\
			\hline
			&\textit{Heavy}&-&-&-&-&-&0.6631&0.6682\\
			\textit{Multiple Haze Level}&\textit{Medium}&-&-&-&-&-&0.6903&0.6925\\
			&\textit{Light}&-&-&-&-&-&0.6962&0.7032\\
			\hline
		\end{tabular}
	\end{center}		
	\label{tab:retrained_joint}
\end{table*}
	
		\begin{table*}
			\caption{The mAP comparison with the Auto-Faster R-CNN scheme on the synthetic hazy sets from PASCAL VOC 2007}
			\small
			\begin{center}
				\begin{tabular}{cccc}
					\hline
					Training Set&Test Set&Auto-Faster R-CNN&JAOD-Faster R-CNN\\
					\hline
					\textit{Heavy}&\textit{Heavy}&0.6573&0.6819\\
					\hline
					\textit{Medium}&\textit{Medium}&0.6815&0.6931\\
					\hline
					\textit{Light}&\textit{Light}&0.6868&0.7004\\
					\hline
				\end{tabular}
			\end{center}		
			\label{tab:auto-faster}
		\end{table*}

\section{Beyond Restoration: Evaluating and Improving Dehazing on Object Detection}
\label{detection}

High-level computer vision tasks, such as object detection and recognition, concern visual semantics and have received tremendous attentions \cite{NIPS2015_5638, yu2016unitbox}. However, the performance of those algorithms may be largely jeopardized by various degradations in practical applications. The conventional approach resorts to a separate image restoration step before feeding into the target task. Recently, \cite{wang2016studying,liu2017image} validates that a joint optimization of the restoration and recognition steps would significantly boost the performance over the traditional two-stage approach. However, previous works \cite{zhang2011close, diamond2017dirty,ACII17} mainly examined the effects and remedies for common degradations such as noise, blur and low resolution, on image classification tasks only. To our best knowledge, there has been no similar work to quantitatively study how the existence of haze would affect high-level vision tasks, and how to alleviate its impact using joint optimization methods. 

We study the problem of object detection in the presence of haze, as an example for how high-level vision tasks can interact with dehazing. We choose the Faster R-CNN model~\cite{NIPS2015_5638} as a strong baseline\footnote{We use the VGG16 model pre-trained based on 20 classes of Pascal VOC 2007 dataset, as provided by the Faster R-CNN authors: \url{https://github.com/rbgirshick/py-faster-rcnn}}, and test on both synthetic and natural hazy images. We then concatenate the AOD-Net model with the Faster R-CNN model, to be jointly optimized as a unified pipeline. General conclusions drawn from our experiments are: as the haze turns heavier, the object detection becomes less reliable. In all haze conditions (light, medium or heavy), our jointly tuned model constantly improves detection, surpassing both naive Faster R-CNN and non-joint approaches. 


\begin{figure*}
	\subfigure[Naive Faster R-CNN]{
		\includegraphics[width=2.4in,height=3in]{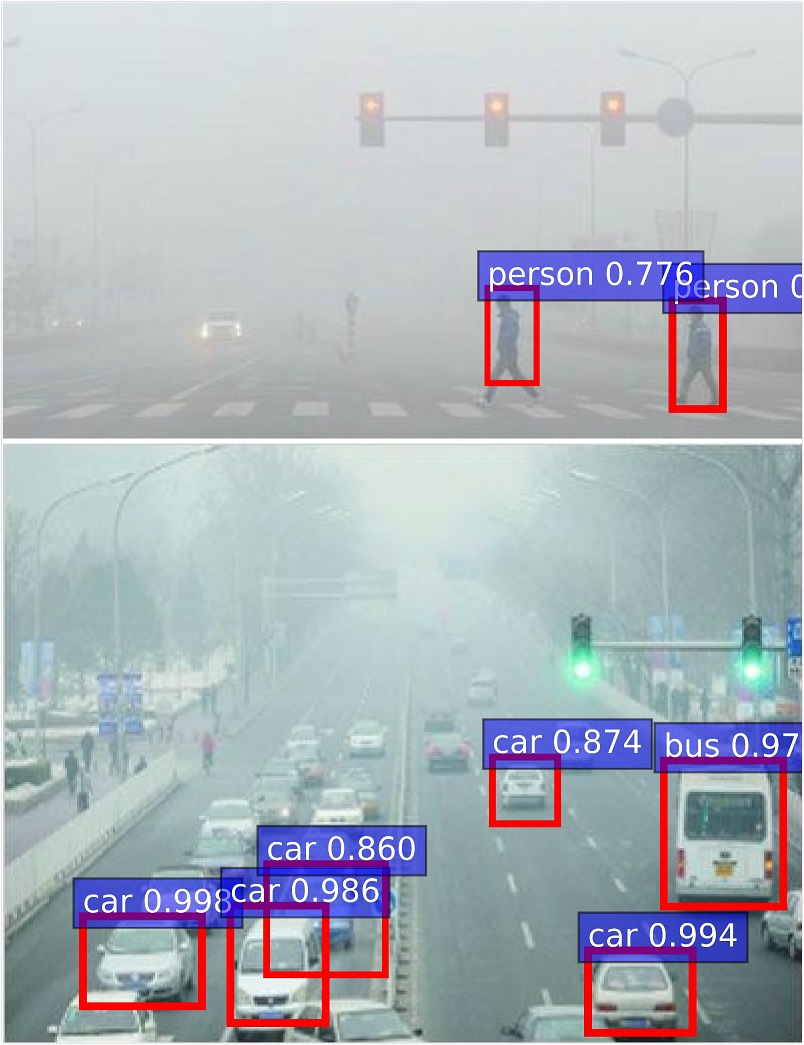}}
	\subfigure[DehazeNet + Faster R-CNN]{
		\includegraphics[width=2.4in,height=3in]{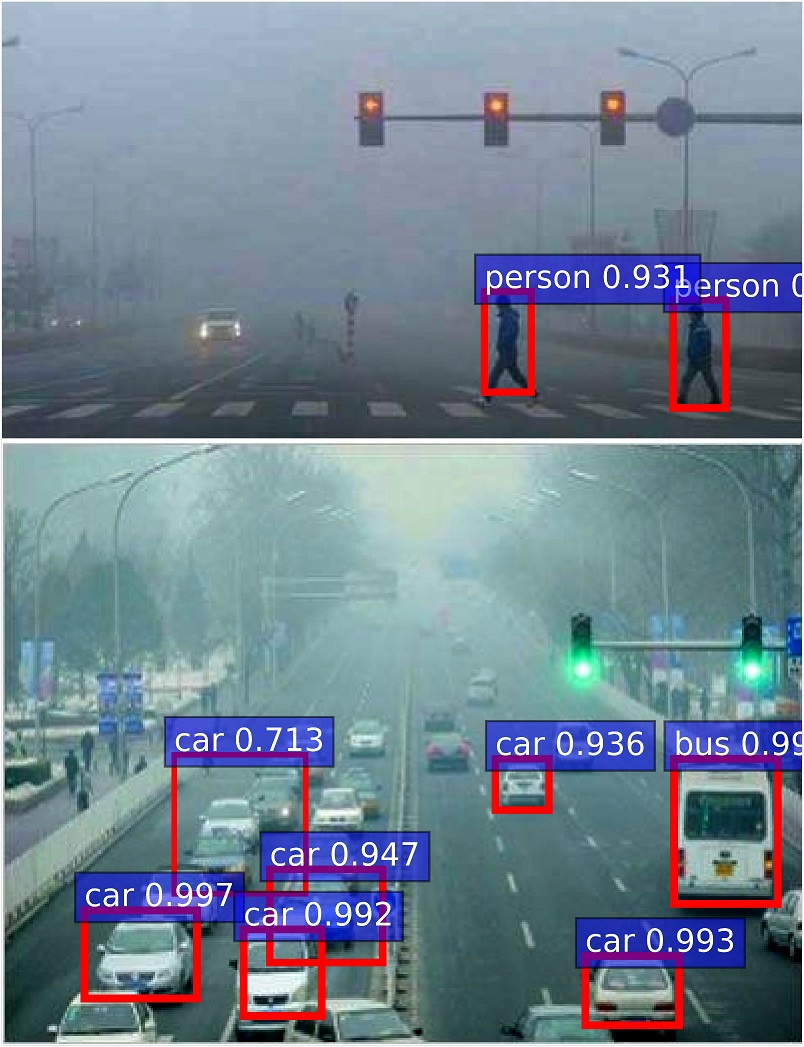}}
	\subfigure[MSCNN + Faster R-CNN]{
		\includegraphics[width=2.4in,height=3in]{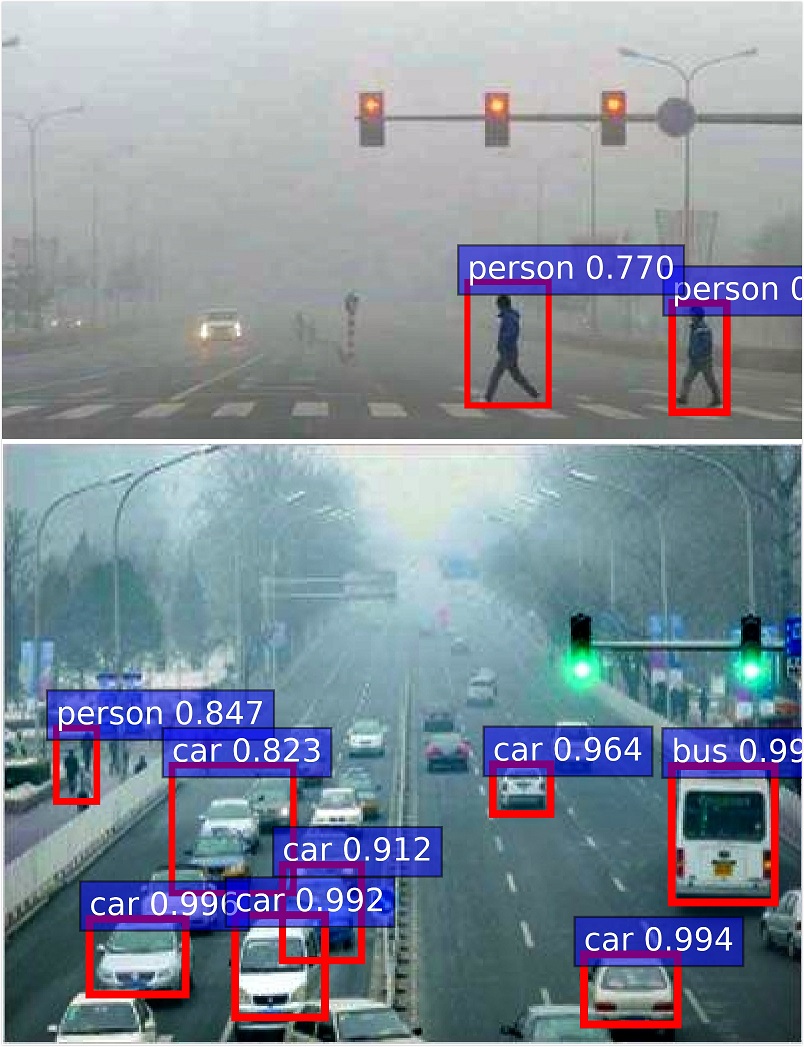}}\\
	\subfigure[AOD-Net + Faster R-CNN]{
		\includegraphics[width=2.4in,height=3in]{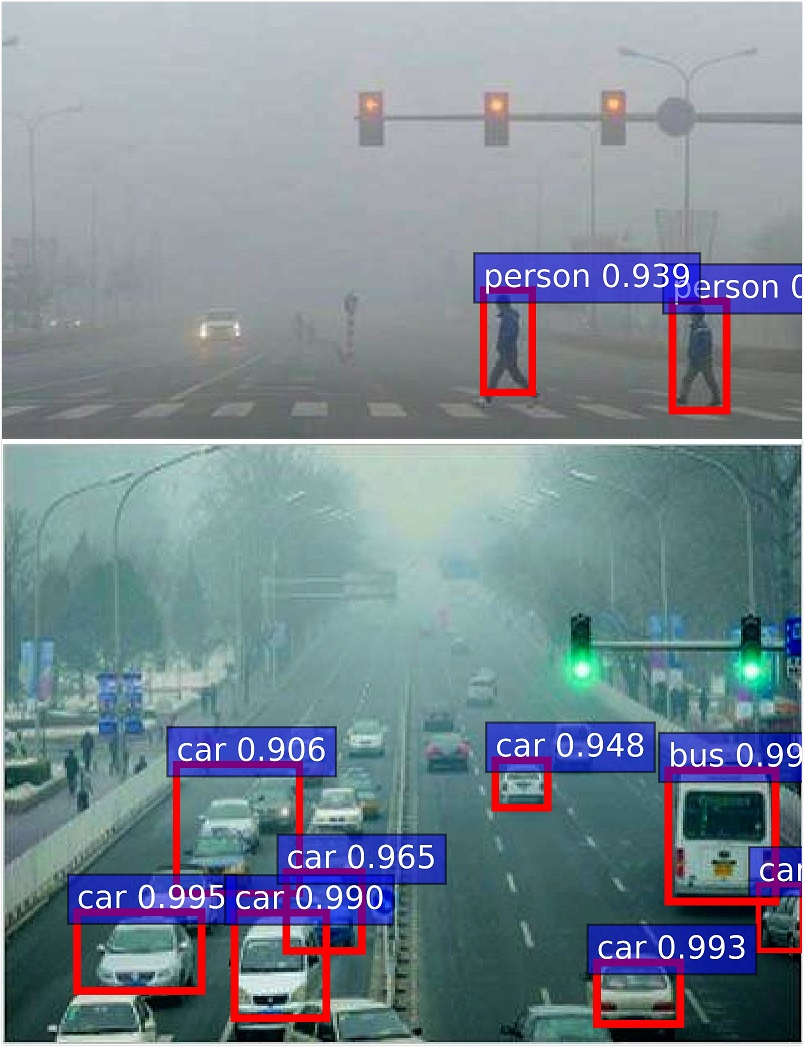}}
	\subfigure[Retrained Faster R-CNN]{
		\includegraphics[width=2.4in,height=3in]{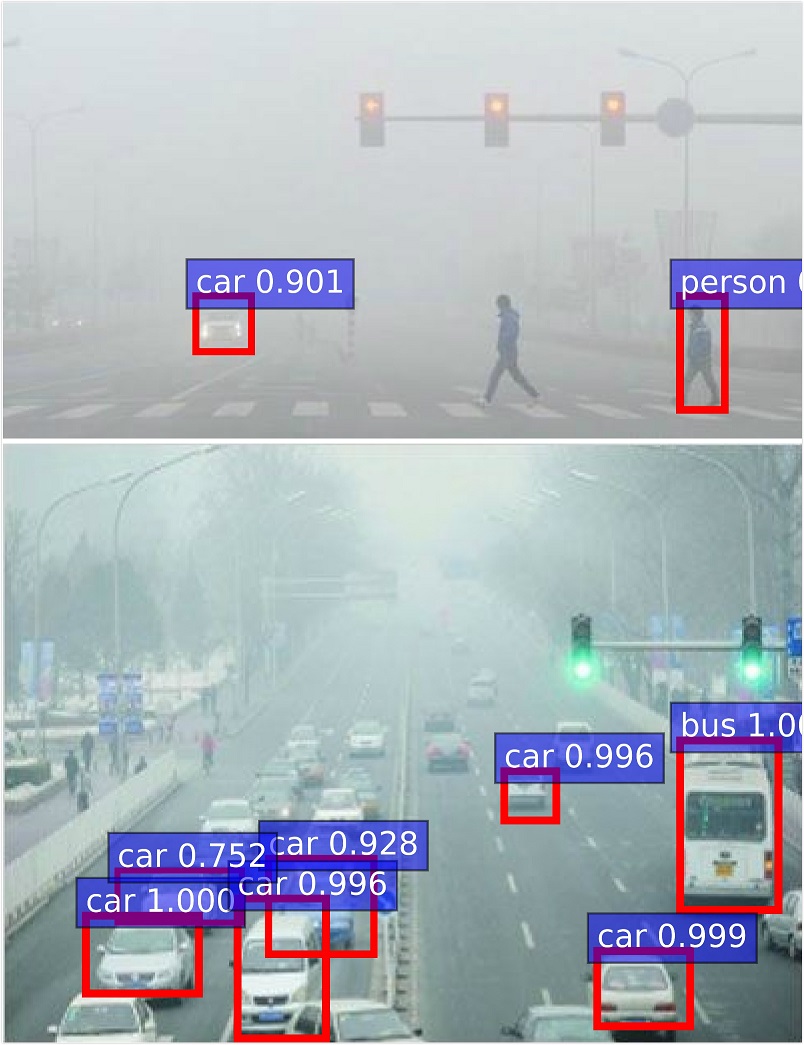}}
	\subfigure[JAOD-Faster R-CNN]{
		\includegraphics[width=2.4in,height=3in]{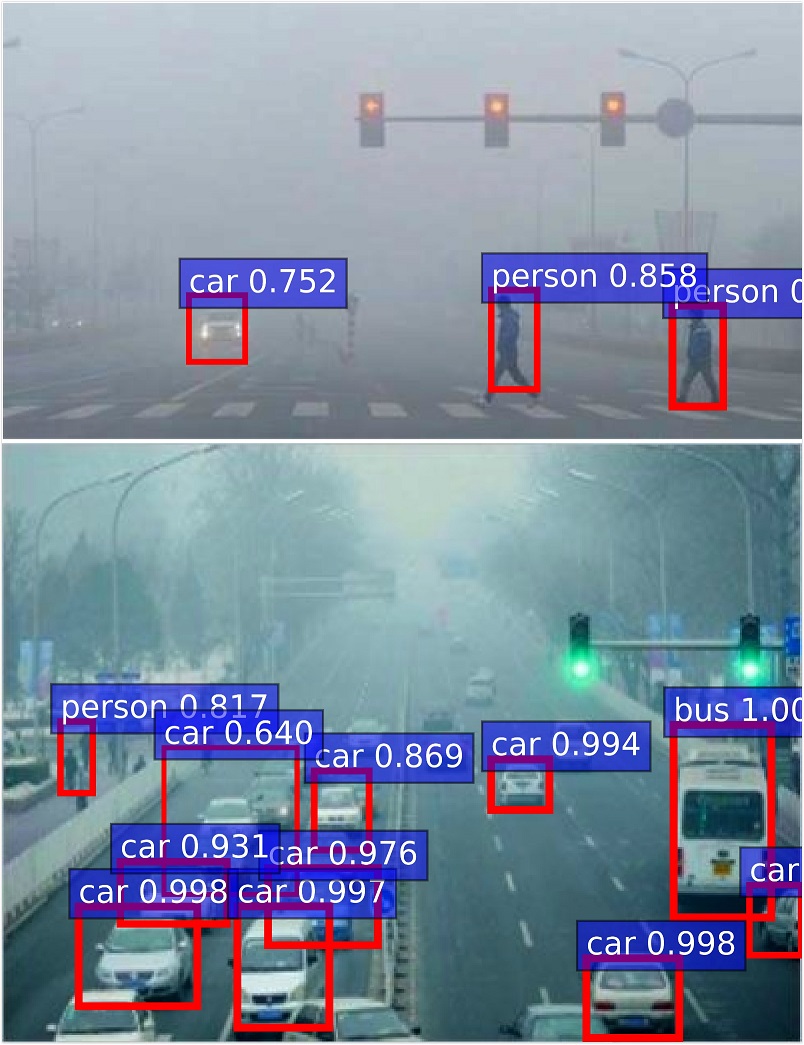}}
	\hspace{0.2in}
	\caption{Object detection results on natural hazy images with a confidence threshold of 0.6, displayed on the dehazed results (except (a) (e) where no dehazing is performed; for (f), we visualize the output of the tuned AOD-Net part as the dehazed image.). For fine-tuned Faster R-CNN and JAOD-Faster R-CNN, we select the model trained with the synthetic \textit{light haze}.
	}
	\label{fig:efficiency_high}
\end{figure*} 

\subsection{Quantitative Results on Pascal-VOC 2007 with Synthetic Haze}
We create three synthetic sets from the Pascal VOC 2007 dataset (referred to as \textit{Groundtruth})~\cite{pascal-voc-2007}: \textit{Heavy Haze} ($ A=1, \beta=0.1$), \textit{Medium Haze} ($ A=1, \beta=0.06$), and \textit{Light Haze} ($ A=1, \beta=0.04$). The depth maps are predicted via the method described in~\cite{liu2016learning}. Each set is split into non-overlapped training set and test set. First we compare five schemes without any network fine-tuning: (1) \textit{naive Faster-RCNN}: directly input with the hazy image using the model pretrained on clean Pascal-VOC 2007; (2) \textit{DehazeNet + Faster R-CNN}: first dehazing using DehazeNet and then Faster R-CNN; (3) \textit{MSCNN + Faster R-CNN}: first dehazing using MSCNN and then Faster R-CNN; (4) \textit{DCP + Faster R-CNN}: first dehazing using DCP and then Faster R-CNN; (5) \textit{AOD-Net + Faster R-CNN}: AOD-Net concatenated with Faster R-CNN, without any joint tuning.

We calculate the mean average precision (mAP) on the three test sets, as shown in the first three rows in Table ~\ref{tab:retrained_joint}. The mAP on the clean Pascal-VOC 2007 test set is 0.6954. We can see that the heavy haze degrades mAP for nearly 0.18. By first dehazing using various dehazing methods before detection, the mAP improves a lot. Among them, \textit{DCP+Faster R-CNN} performs the best with 21.57\% improvement in the heavy haze. Without any joint tuning, \textit{AOD-Net+Faster R-CNN} performs comparable to \textit{MSCNN+Faster R-CNN}, and appears a bit worse than \textit{DCP+Faster R-CNN}. 


Owing to our all-in-one design, the pipeline of AOD-Net+Faster R-CNN can be jointly optimized from end to end for improved object detection performance on hazy images. We tune \textit{AOD-Net+Faster R-CNN} for the three hazy training sets separately, and call this tuned version \textit{JAOD-Faster R-CNN}. We use a learning rate of 0.0001 for first 35, 000 iterations, and 0.00001 for next 65, 000 iterations,  both with a momentum of 0.9 and a weight decay of 0.0005. As a result of such joint tuning, the mAP increases from 0.5794 to 0.6819 for the heavy haze case, which shows a major strength of such end-to-end optimization and the value of our unique design. For comparison, we also re-train Faster R-CNN on the hazy data sets as a comparison. We use a learning rate of 0.0001 for fine-tuning a pre-trained Faster R-CNN (trained on clean natural images). 
After retraining to be adapted to the hazy dataset, the mAP of Retrained Faster R-CNN increases from 0.5155 to 0.6756 under heavy haze, while still being consistently worse than \textit{JAOD-Faster R-CNN}. 

\begin{figure*}
		\includegraphics[width=.5\linewidth - 0.25mm]{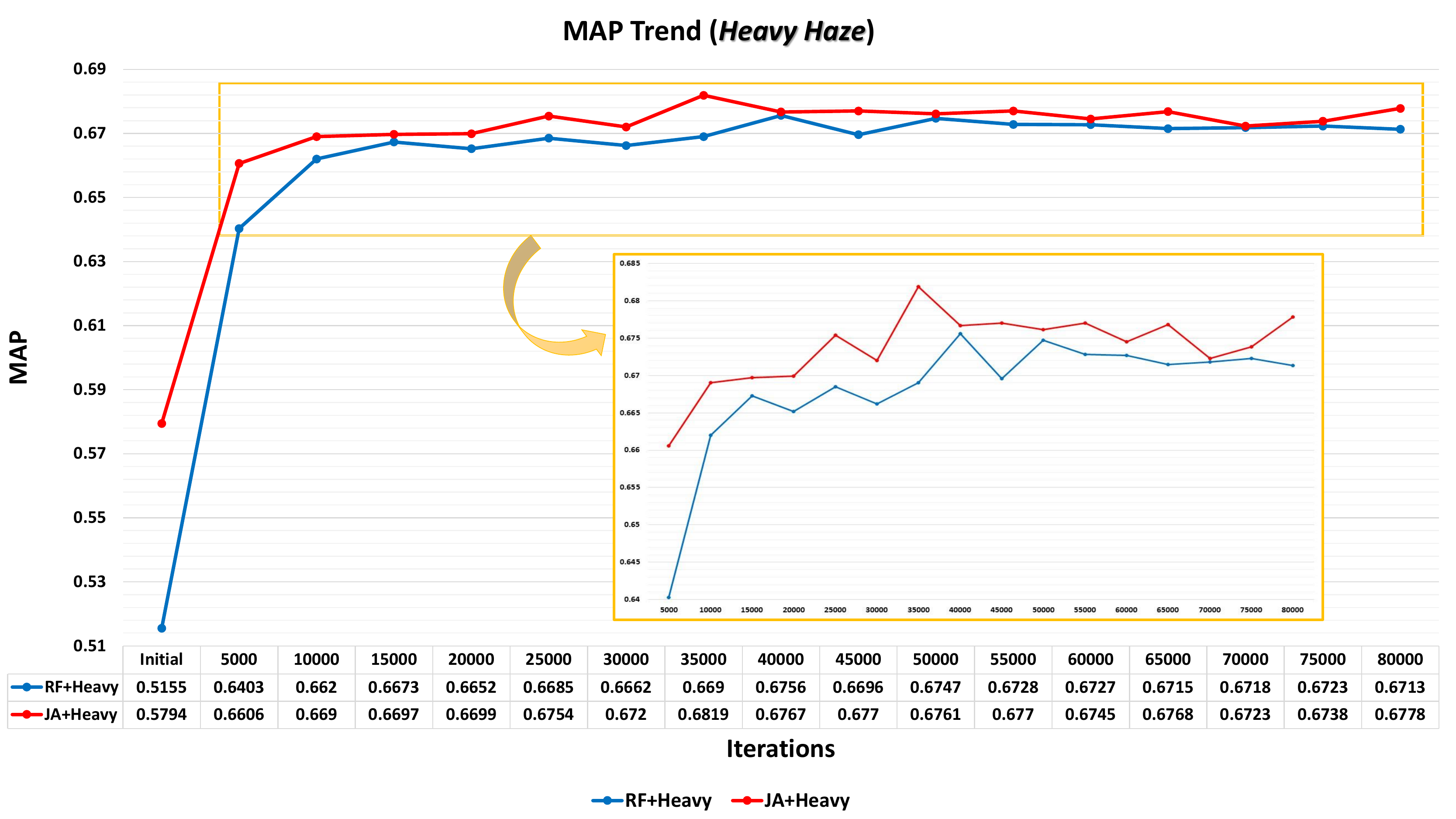}\hfill
		\includegraphics[width=.5\linewidth - 0.25mm]{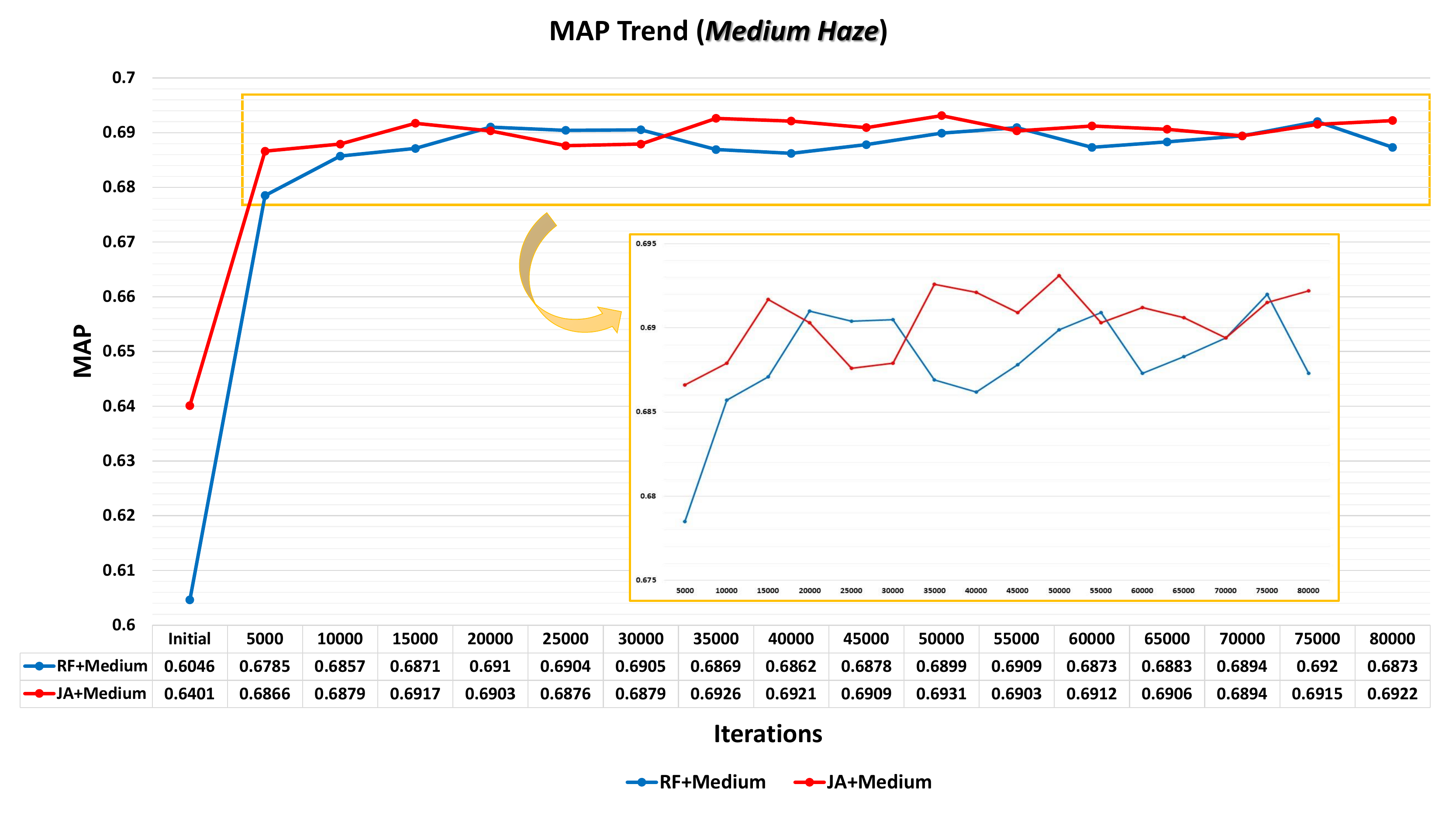}\\[0.5mm]
		\includegraphics[width=.5\linewidth - 0.25mm]{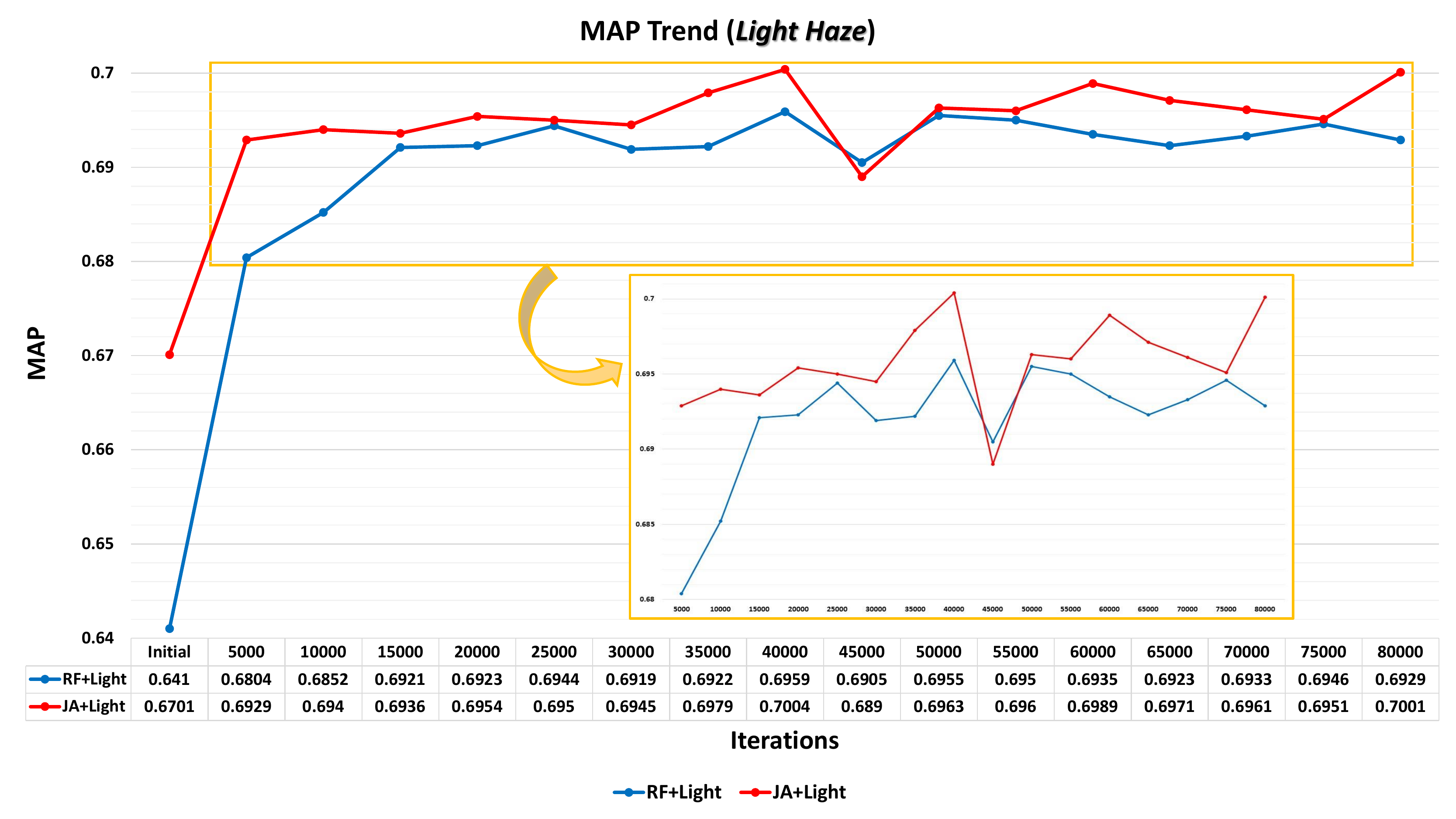}\hfill
		\includegraphics[width=.5\linewidth - 0.25mm]{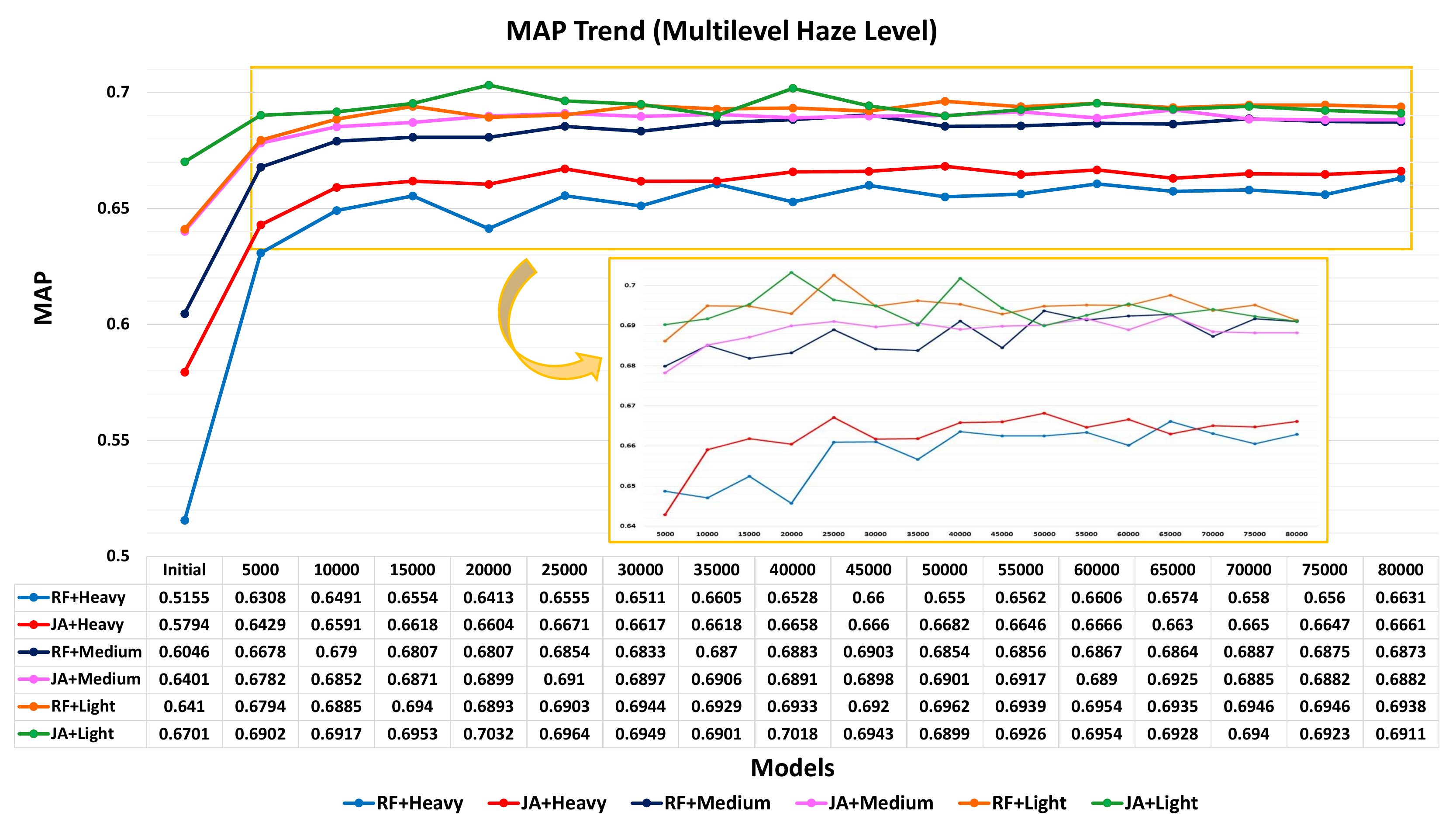}
		\hspace{0.2in}
		\caption{The mAP over the training process at every 5, 000 iterations, for four training datasets. (a) Training Set: \textit{Heavy Haze}(Heavy). (b) Training Set: \textit{Medium Haze}(Medium). (c) Training Set: \textit{Light Haze}(Light). (d) Training Set: \textit{Multiple Haze Levels}. (RF: Retrained Faster R-CNN, JA: JAOD-Faster R-CNN)}
		\label{fig:MAP_trend}
	\end{figure*}

Furthermore, since it is practically desirable to obtain one unified model that works for arbitrary haze levels, we generate a training set that includes various haze levels with $ \beta $ randomly sampled from $[0,0.1]$. We re-tune and evaluate \textit{JAOD-Faster R-CNN} and \textit{Retrained Faster R-CNN} on this training set, whose results are  compared in the last row of Table ~\ref{tab:retrained_joint}. Although both perform slightly inferior to their ``dedicated'' counterparts trained and applied for a specific haze level, they perform consistently well in all three haze levels, and \textit{JAOD-Faster R-CNN} again outperforms \textit{Retrained Faster R-CNN}. Figure~\ref{fig:MAP_trend} plots the mAP comparisons at every 5, 000 iterations, between the JAOD-Faster R-CNN and Retrained Faster R-CNN schemes, under various haze conditions. 

\subsection{Visualized Results on Natural Hazy Images} Figure~\ref{fig:efficiency_high} displays a visual comparison of object detection results on web-source natural hazy images. Six approaches are compared: (1) \textit{naive Faster-RCNN}; (2) \textit{DehazeNet + Faster R-CNN}; (3) \textit{MSCNN + Faster R-CNN}; (4) \textit{AOD-Net + Faster R-CNN}; (5) \textit{Fine-tuned Faster R-CNN}. (6) \textit{JAOD-Faster R-CNN}. 
We observe that haze can cause missing detections, inaccurate localizations and unconfident category recognitions for Faster R-CNN. While \textit{AOD-Net + Faster R-CNN} already shows visible advantages over \textit{naive Faster-RCNN}, the performance is further dramatically improved in \textit{JAOD-Faster R-CNN} results, which surpasses all other alternatives visibly.
	
Note that AOD-Net + Faster R-CNN benefits from joint optimization in two-folds: the AOD-Net itself jointly estimates all parameters in one, and the entire pipeline jointly tunes the low-level (dehazing) and high-level (detection and recognition) tasks from end to end. The end-to-end pipeline tuning is uniquely made possible by AOD-Net, which is designed to be the only all-in-one dehazing model so far.

	
	
	%
	%

\subsection{Who is actually helping: the task-specific dehazing network or just adding more parameters?}

While JAOD-Faster R-CNN is arguably the best performer as shown above, one question may arise naturally: is it just a result of the fact that AOD-Faster R-CNN uses more parameters than (Retrained) Faster R-CNN? In this section, we show that adding extra layers and parameters, without a task-specific design for dehazing, does not necessarily improve the performance of object detection in the haze.


We designed a new baseline, named \textit{Auto-Faster R-CNN}, that replaces the AOD-Net part in JAOD-Faster R-CNN with a plain convolutional auto-encoder. The auto-encoder has exactly the same amount of parameters with AOD-Net, consisting of five convolutional layers with its structure resembling the $K$-estimation module. We pre-train the auto-encoder for the dehazing task, using the same training protocol and dataset with AOD-Net, and the concatenate it with Faster R-CNN for end-to-end tuning. As observed in Table~\ref{tab:auto-faster}, the performance of Auto-Faster R-CNN is not on par with AOD-Faster R-CNN, and shows even worse than Fine-tuned Faster R-CNN. Recall that \cite{NIPS2015_5638} verified that directly adding extra layers to Faster R-CNN did not necessarily improve the performance of object detection in general clean images. Our conclusion is its consistent counterpart in the hazy case.



Besides, it should be noted that although \textit{JAOD-Faster R-CNN} appends AOD-Net before Faster R-CNN, the complexity is not increased much, thanks to the light-weight design of AOD-Net. The per-image average running time is 0.166s for \textit{(Re-trained) Faster R-CNN}, and 0.192s for \textit{JAOD-Faster R-CNN}, using the NVIDIA GeForce GTX TITAN X GPU.

%
%

\section{Discussion and Conclusions}
	
The paper proposes AOD-Net, an all-in-one pipeline that directly reconstructs haze-free images via an end-to-end CNN. We compare AOD-Net with a variety of state-of-the-art methods, on both synthetic and natural haze images, using both objective (PSNR, SSIM) and subjective criteria. Extensive experimental results confirm the superiority, robustness, and efficiency of AOD-Net. Moreover, we also present the first-of-its-kind study, on how AOD-Net can boost the object detection and recognition performance on natural hazy images, by joint pipeline optimization. It can be observed that our jointly tuned model constantly improves detection in the presence of haze, surpassing both naive Faster R-CNN and non-joint approaches. Still, as has been mentioned above, the dehazing technique is highly correlated with the depth estimation from images, and there is room for improving the performance of AOD-Net, by incorporating the depth prior knowledge or an elaborate depth estimation module.


%





\ifCLASSOPTIONcaptionsoff
  \newpage
\fi



%

\bibliographystyle{ieee}
\bibliography{egbib}

\end{document}